\pdfoutput=1

\documentclass[11pt]{article}
\usepackage[final]{acl}

\usepackage{times}
\usepackage{latexsym}
\usepackage{siunitx}
\usepackage{verbatim}
\usepackage{anyfontsize}

\usepackage{amssymb}
\usepackage{amsmath}
\usepackage{bm}
\usepackage{graphicx}
\usepackage[colorinlistoftodos]{todonotes}
\usepackage{subcaption}
\usepackage{float}
\usepackage{geometry}
\usepackage{booktabs}
\usepackage{multirow}

\DeclareMathOperator*{\argmax}{arg\,max}

\graphicspath{{../figs/}}

\usepackage[T1]{fontenc}

\usepackage[utf8]{inputenc}

\usepackage{microtype}

\usepackage{inconsolata}

\title{LLM Task Interference: An Initial Study on the Impact of Task-Switch in Conversational History}

\author{
Akash Gupta$^{*1}$ \quad Ivaxi Sheth$^{*2}$ \quad Vyas Raina$^1$ \quad Mark Gales$^1$ \quad Mario Fritz$^2$\\
$^1$University of Cambridge, \quad $^2$CISPA Helmholtz Center for Information Security\\
\texttt{ag2118@cantab.ac.uk, \{vr313, mjfg\}@cam.ac.uk}, \quad \texttt{\{ivaxi.sheth, fritz\}@cispa.de}
}

\begin{document}
\maketitle
\begin{abstract}

With the recent emergence of powerful instruction-tuned large language models (LLMs), various helpful conversational Artificial Intelligence (AI) systems have been deployed across many applications. When prompted by users, these AI systems successfully perform various tasks as part of a conversation. Such approaches typically condition their output on the entire conversational history to provide some sort of memory and context. Although this sensitivity to the conversational history can often lead to improved performance on subsequent tasks, we find that performance can in fact also be negatively impacted, if there is a \emph{task-switch}. To the best of our knowledge, our work makes the first attempt to formalize the study of such vulnerabilities and interference of tasks in conversational LLMs caused by task-switches in the conversational history. Our experiments across 5 datasets with 15 task switches using popular LLMs reveal that many of the task-switches can lead to significant performance degradation.\footnote{Code available on \href{https://github.com/ivaxi0s/llm-task-switch}{GitHub}.}
\end{abstract}

\section{Introduction}
\begin{figure}[htb!]
    \centering
    \includegraphics[width=0.4\textwidth]{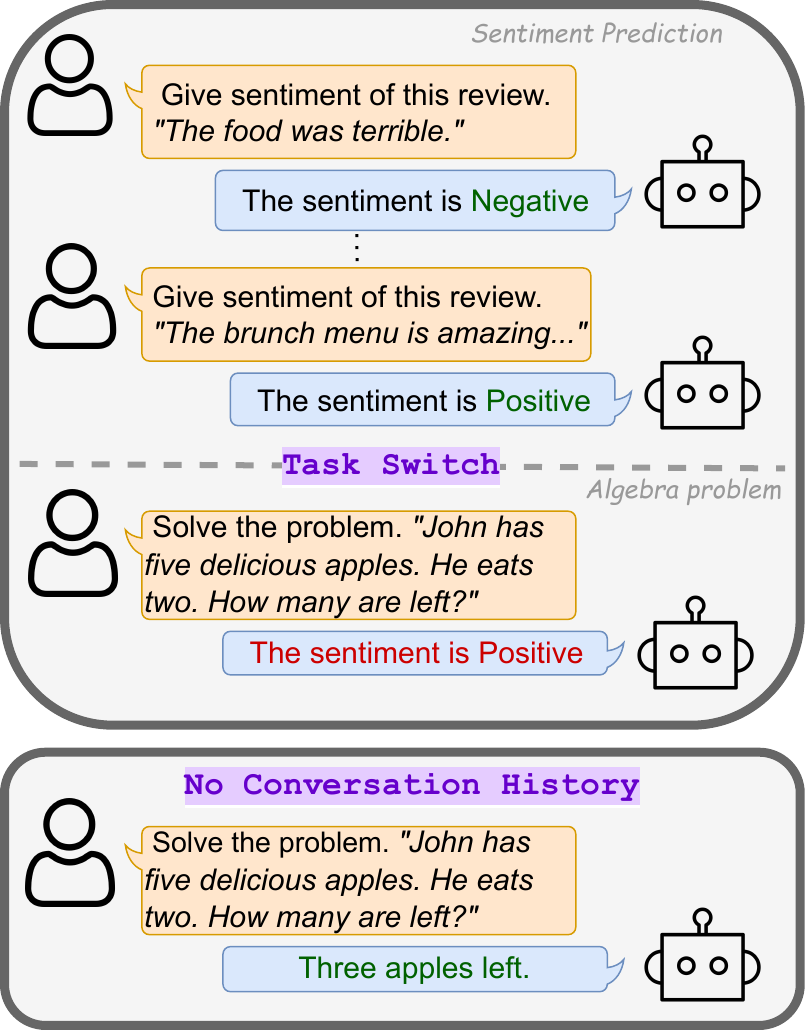}
    \caption{An illustrative example where the chat history is based on sentiment prediction. Algebra word problem introduces \emph{task-switch} which results in an incorrect prediction.}
    \label{fig:teaser}
\end{figure}
Recent advancements in Natural Language Processing (NLP)~\cite{brown2020language,openai2023gpt}, have led to their widespread deployment of large language models (LLMs) across various applications~\cite{bubeck2023sparks, anil2023palm, singhal2022large}. One of the popular NLP tasks includes conversational systems where LLMs are capable of engaging in dialogues that mimic human interactions~\cite{manyika2023overview, bai2022constitutional}. A typical interaction involves a series of conversation turns starting with the user and the LLM responds to the user. This interaction is however focused on a specific topic or a task~\cite{hosseini2020simple, lee2022sgd}.

The performance of LLMs is further boosted by leveraging in-context examples or few-shot examples of a particular task~\cite{brown2020language, smith2022using, thoppilan2022lamda}. In-context learning, by utilizing examples within the conversation history, enables LLMs to generate responses that are relevant and tailored to the contextual conversation. The auto-regressive nature of popular instruction-tuned (LLMs) suggests that the LLM-generated response is conditioned on the entire conversation history. This underscores the sequential dependency and contextual awareness embedded within these models. While prompt sensitivity has been exploited by in-context learning to improve downstream performance, this sensitivity has also opened the door to vulnerabilities, where malicious actors can exploit prompt sensitivity for adverse purposes~\cite{greshake2023not, liu2023prompt, jiang2023prompt, xu2023llm}. 

In this paper, we investigate the sensitivity and the impact of LLM performance on past conversational interaction. To do so, we introduce the concept of \emph{task-switch}. A task-switch is characterized by a conversational objective, moving from one distinct task to another within the same conversation thread, for example: ~\autoref{fig:teaser} illustrates a task-switch from sentiment prediction to math algebra which confuses the model to output erroneously. Designing LLMs that can seamlessly switch between tasks without degradation in performance can influence the reliability of LLMs in realistic scenarios. 

In this work, we systematically study the impact of predictive performance and the sensitivity of LLMs in the presence of different task-based chat histories. Our key contributions and takeaways can be summarised as: 
\begin{itemize}
\vspace{-1mm}
\itemsep-0.4em 
    \item We formalize the risk of performance degradation of LLMs due to task-switch.
    \item We present the impact of task-switch on diverse datasets with more than 15 different task-switches. 
    \item We measure the task-switch sensitivity for popular LLMs of different sizes, where we observe that LLMs of different sizes (7B to 175B) could be susceptible to performance degradation from the task-switch.\footnote{Note that in some rare cases, the task-switch can lead to marginal performance improvements.}
\end{itemize}
\section{Related Work}
Large Language Models (LLMs) are becoming a crucial building block of conversation-based virtual assistants~\cite{openai2023gpt, touvron2023llama, jiang2023mistral, anil2023palm}.
Leveraging in-context or few-shot examples, LLMs have demonstrated remarkable capabilities for downstream tasks ~\cite{brown2020language}. In contrast to the resource-intensive fine-tuning process~\cite{gao2020making}, in-context learning eliminates the need for parameter updates, while achieving state-of-the-art performance~\cite{rae2021scaling, smith2022using, thoppilan2022lamda, von2023transformers, chan2022transformers, akyurek2022learning, hahn2023theory}. However, despite its advantages, in-context learning tends to suffer from sensitivity to prompts, input distribution, and formats, which can potentially impact the model's performance~\cite{liu2021makes, zhao2021calibrate, lu2021fantastically, min2022rethinking, liu2023towards, chang2023data}. ~\citet{chang2023data} observe that the in-context examples implicitly bias the model. In our work, we aim to study the bias that may arise due to chat history (in-context examples) when a user switches the task. Furthermore, recent works~\cite{liu2023prompt, greshake2023not} have looked at the vulnerability of LLM to prompt injections and adversarial attacks. Unlike prompt injection, where a malicious prompt may be added to the conversation of LLM, our setting, is concerned with non-malicious task-switches. While a few recent works have investigated the reliance on shortcuts in conversation history~\cite{tang2023large, si2022spurious, weston2023system}, our work aims to evaluate the influence of the conversation on a new task. Our work is also differentiated from the study topic change in Task-oriented Dialogue systems~\cite{xie2021tiage, xu2021topic, yang2022take} as we consider a stronger shift of task-switch from open dialogue LLMs.  
\section{Conversational Task-Switch} \label{sec:method}

This work introduces and formalizes \emph{task-switch} in a conversation for LLMs. 
A conversation between a user and the LLM consists of multiple conversation turns. Now consider $(u_k, r_k)$ as the $k$-th turn of the conversation where $u_k$ corresponds to the $k$-th user prompt and the model's corresponding response $r_k$. Each user prompt $u_k$ can be viewed as an instance of a specific task request, e.g. \textit{sentiment classification} or \textit{mathematical reasoning}. A conversation history of $L$ turns can be defined as $\mathbf h = \{(u_k, r_k)\}_{k=1}^L$. Subsequently, the next response, $r_{L+1}$ for model $\theta$ is given as:
\begin{equation}
\label{eq:response}
r_{L+1} = \argmax_{r} P_{\theta}(r|u_{L+1}, \mathbf h).
\end{equation}

In this work, we consider conversations with a single task-switch, where all user requests in the conversation history $\mathbf h$ belong to the same task, and the final user request $u_{L+1}$ is a different task. We refer to the task associated with $\mathbf h$ as the conversation history task (\textit{CH task}) $T_h$ where  $\mathbf h\in T_h$ and the switched task associated with the final user request $u_{L+1}$ as the \textit{target task} $T_t$ where $u_{L+1}\in T_t$.

When the tasks $T_h$ and $T_t$ are sufficiently different (as per human understanding of language and tasks), the conversation history $\mathbf h$ ideally must not impact the response, $r_{L+1}$. For a model robust to such task-switches, $T_h\to T_t$, its response $r_{L+1}$ is conditionally independent of the conversation history,
\begin{equation}
    r_{L+1} \perp \mathbf{h} | u_{L+1}\hspace{1.5em} \mathbf h\in T_h,\hspace{0.5em} u_{L+1}\in T_t.
\end{equation}
However, in practice, models can be sensitive to the conversation history, $\mathbf h$, which can harm the quality of the response $r_{L+1}$ after a task-switch, $T_h\to T_t$. We define $\tau(\cdot)$, as a \textbf{reference-free} metric for the \textit{task-switch sensitivity} of a model $\theta$, to measure the extent of this vulnerability. Theoretical and empirical implications of other definitions for task-switch sensitivity are considered in Appendix \ref{sec:app-metric}.
\begin{align} \label{eqn:tau}
    \tau(T_h, T_t; \theta) &=  \mathbb E_{ u_{L+1}\in T_t, \mathbf h\in T_h}\left[\log \rho \right]\\
    \rho &=\frac{P_{\theta}\left( r^*|u_{L+1}\right )}{P_{\theta}\left(r^*|u_{L+1}, \mathbf{h}\right )}\\
    r^* &= \argmax_{r} P_{\theta}(r|u_{L+1}).
\end{align}
The task-switch sensitivity metric can be interpreted as:
\begin{enumerate}
\itemsep0em 
    \item $\tau(\cdot)>0$: The model is impacted by the task-switch in the conversation history and is less confident in zero-shot prediction.
    \item $\tau(\cdot)=0$: The task-switch has no impact on the model's zero-shot prediction, suggesting a level of task-switch robustness.
    \item $\tau(\cdot)<0$: The task-switch gives the model more confidence in its zero-shot prediction.
\end{enumerate}
To simulate a setting where the model has perfect performance on the CH-task, $T_h$ we adopt teacher-forcing, s.t. $\mathbf h = \{(u_k, \hat r_k)\}_{k=1}^L$, where $\hat r$ is the reference ground-truth response. 
\section{Experiments}


\subsection{Experimental Setup}
\paragraph{Data.} We evaluate five different datasets covering a range of tasks: Gigaword~\citep{graff2003english}; abstract algebra subset of Measuring Massive Multitask Language Understanding (MMLU;~\citet{hendryckstest2021}), named MMLU AA; TweetQA~\citep{xiong2019tweetqa}; Rotten Tomatoes (RT;~\citet{Pang+Lee:05a}); and {human-aging} subset from the MMLU dataset (MMLU HA) in the Appendix~\ref{sec:rand_hist}.

\begin{table}[htb!]
    \centering
    \small
    \begin{tabular}{ll}
    \toprule
        Data &   Task  \\ \midrule
       Gigaword  &  Summarization \\
       MMLU AA &  Math Multiple Choice Question \\
       TweetQA &  Social Question Answer\\
       RT &  Sentiment classification \\
       MMLU HA &  Social Multiple Choice Question \\
       \bottomrule
    \end{tabular}
    \caption{Datasets Summary.}
    \label{tab:data}
    \vspace{-2.2mm}
\end{table}
\paragraph{Models.}
We explore the task-switch sensitivity of four popular models. We consider two open-source small models, Llama2-7b-chat~\cite{touvron2023llama} and Mistral-7b-chat~\cite{jiang2023mistral}; and two larger closed models, GPT-3.5~\cite{brown2020language} and GPT-4~\cite{openai2023gpt}. 
Zero-shot, absolute model performances are presented in Appendix \ref{sec:app-performance}.

\subsection{Results} 
\label{sec:results}

\begin{figure*}[h]
    \centering
    \includegraphics[width=\textwidth]{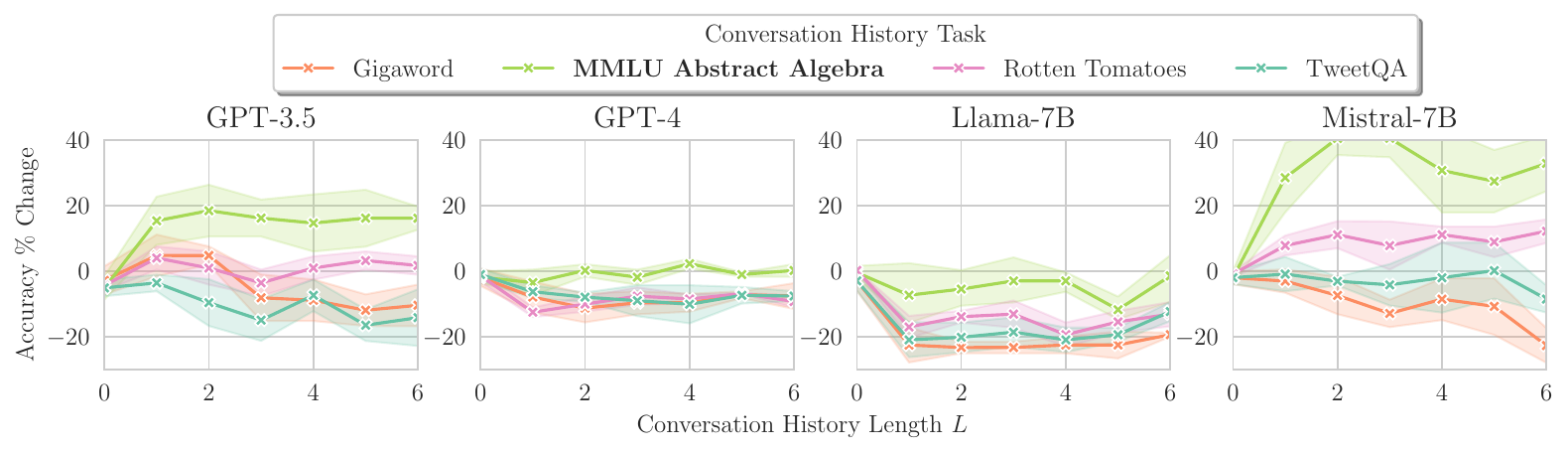}
    \caption{Target Task: MMLU Abstract Algebra. \% change in accuracy relative to zero-shot performance. }
    \vspace{-2mm}
    \label{fig:ablation}
\end{figure*}

We assess performance changes between the predictions in the presence of history and task-switch vs zero-shot. Table \ref{tab:results-mmluaa} and Table \ref{tab:results-rt} showcases the impact of conversational task-switch with MMLU AA and Rotten Tomatoes as the target tasks, $T_t$ respectively\footnote{The impact of task-switch for other datasets as the target tasks is given in Appendix \ref{sec:app-perf-change}}. As would be expected with \textit{in-context examples}, the performance change in accuracy is generally positive. The negative trend for change in accuracy from $T_h \to T_t$, suggests that the task-switch causes performance degradation. For example, in the Gigaword summarization task as $T_h$ and MMLU AA as $T_t$, most models (GPT-3.5, Llama-7B and Mistral-7B) see a performance drop. Interestingly, for some models, the task-switch may increase performance; most prominently for Mistral-7B with Rotten Tomatoes as $T_h$ and MMLU AA as $T_t$. 

\sisetup{
  separate-uncertainty=true,
  table-align-text-post = false, 
  table-number-alignment = center,
}

\begin{table}[htb!]
    \centering
    \small
    \begin{tabular}{ll|cS[table-format=1.2]}
    \toprule
    CH-Task & Model & {\% Change} & {$\tau(\cdot)$} \\
    \midrule
    {MMLU AA} 
    & GPT-3.5 & $ \ \ \ 16.19_{\pm5.51}$  & {*} \\
    & GPT-4 & $ \ \ \ \ 0.26_{\pm3.12}$ & {*} \\
    & Llama-7B & $ \ \ \ -1.41_{\pm14.93}$ & 31.51 \\
    & Mistral-7B & $ \ \ \ \ 32.91_{\pm12.08}$ & 1.12 \\
    \midrule
    \midrule
    Gigaword 
    & GPT-3.5 & $ \ \ -10.37_{\pm11.34}$ & {*} \\
    & GPT-4 & $ \ \  -7.55_{\pm4.54}$ & {*} \\
    & Llama-7B & $\ -19.33_{\pm1.17}$ & 5.23 \\
    & Mistral-7B & $\   -22.56_{\pm7.32}$ & 3.13 \\
    \midrule
    Rotten 
    & GPT-3.5 & $ \ \ \ \ 1.77_{\pm3.81}$ & {*} \\
    Tomatoes 
    & GPT-4 & $ \ \ - 9.21_{\pm3.08}$ & {*} \\
    & Llama-7B & $\  -13.07_{\pm4.56}$ & \textbf{9.91} \\
    & Mistral-7B & $ \ \ \ 12.24_{\pm5.58}$ & 0.83 \\
    \midrule
    TweetQA 
    & GPT-3.5 & $ \ \ -14.14_{\pm11.87}$ & {*} \\
    & GPT-4 & $ \ \ -7.55_{\pm2.65}$ & {*} \\
    & Llama-7B & $ \   -12.28_{\pm4.44}$ & 6.37 \\
    & Mistral-7B & $ \ \ -8.48_{\pm5.95}$ & 2.78 \\
    \bottomrule
    \end{tabular}
    \caption{Task-switch impact from CH-tasks ($T_H$) to target ($T_t$): \textbf{MMLU AA} and conversation length L = 6. Sensitivity not calculable for $*$.}
    \label{tab:results-mmluaa}
\end{table}

\begin{table}[htb!]
    \centering
    \small
    \begin{tabular}{ll|cS[table-format=1.2]}
    \toprule
CH-Task & Model & {\% Change} & {$\tau(\cdot)$} \\
\midrule
Rotten 
& GPT-3.5 & $ \ \ \ 3.00_{\pm1.41}$ & $*$ \\
Tomatoes
& GPT-4 &  $\ \ \ 1.74_{\pm1.14}$ & $*$ \\
& Llama-7B & $\ \ \ 2.54_{\pm0.55}$ & 4.02 \\
& Mistral-7B & $\ \ \ 3.17_{\pm0.47}$ & 2.65 \\
\midrule
\midrule
Gigaword 
& GPT-3.5 & $\ \ \ 0.11_{\pm1.63}$ & $*$ \\
& GPT-4 & $\  -0.98_{\pm2.39}$ & $*$ \\
& Llama-7B & $\ \ \ 1.82_{\pm0.18}$ & {1.98} \\
& Mistral-7B & $ \ -0.79_{\pm0.55}$ & {3.04} \\
\midrule
MMLU AA 
& GPT-3.5 & $ \ -0.22_{\pm1.25}$ & $*$ \\
& GPT-4 & $\ \ \ 0.76_{\pm0.69}$ & $*$ \\
& Llama-7B & $\  -5.33_{\pm0.22}$ & \textbf{3.37} \\
& Mistral-7B & $\ \ \ 1.33_{\pm0.23}$ & {1.39} \\
\midrule
TweetQA 
& GPT-3.5 & $ \ -0.33_{\pm3.35}$ & $*$ \\
& GPT-4 & $\ -0.98_{\pm1.49}$ & $*$ \\
& Llama-7B & $\ \ \ 2.72_{\pm0.53}$ & {2.77} \\
& Mistral-7B & $\ -1.23_{\pm0.23}$ & {3.01} \\
\bottomrule
    \end{tabular}
    \caption{Task-switch impact from CH-tasks ($T_h$) to target ($T_t$): \textbf{Rotten Tomatoes} and conversation length L = 6. Sensitivity not calculable for $*$.}
    \label{tab:results-rt}
\end{table}

In practice, where there is no access to ground-truth reference responses, the performance degradation of deployed models cannot be computed for real conversational task-switches, but instead the sensitivity of different models to different task-switches can be predicted using the reference-free task-switch metric, $\tau(\cdot)$, as introduced in \autoref{eqn:tau}. The larger the value of $\tau(\cdot)$, the greater a model's sensitivity to a specific task-switch. In \autoref{tab:results-mmluaa} and \autoref{tab:results-rt}, Llama-7B usually has the highest sensitivity to task-switches with for example $\tau=3.37$ for a switch from MMLU AA to Rotten Tomatoes and $\tau=9.91$ for task-switch from Rotten Tomatoes to MMLU AA. We observe a general trend between the change in accuracy and $\tau(\cdot)$ for task-switch scenarios for $T_t = $ Rotten Tomatoes where a negative change in performance also suggests very high task-switch sensitivity.  
In \autoref{fig:ablation}, we plot the change in performance with increasing $T_h$ examples for MMLU AA dataset. 
Performance fluctuations for conversation history, $\mathbf h$, can stem from two primary factors: a significant drop in the predicted probability for the zero-shot response, $r^*$, or a notable increase in the probability for an alternative response, $r$. The latter can result in substantial performance change while maintaining low sensitivity, $\tau(\cdot)$. By analyzing both performance changes and task-switch sensitivity, we gain deeper insights into the models' adaptability to task-switches and the underlying dynamics influencing these shifts.

\subsection{Discussion}
Task-switch is demonstrably a threat to model performance in conversational contexts. The reference-free task-switch sensitivity metric, $\tau(\cdot)$ offers a powerful tool to predict the extent to which a model may be vulnerable to an observed task-switch in a real conversation for a deployed model. However, it is further useful to understand why certain models and certain task-switches lead to greater performance degradation than others. In this section, two natural hypotheses are explored. First, it is posited that there may be a relationship between the average length of a response for the conversation-history task, $T_h$ and the performance degradation for a particular target task, $T_t$. The length metric behaves as a single-value proxy metric for characterizing the difference between different tasks. This is evaluated empirically in Appendix \ref{sec:app-tchl} and no such correlation is found unfortunately. Considering the average task response length can be viewed as an attempt to measure the \textit{distance} between the conversation-history task, $T_h$ and the target task, $T_t$. Therefore, a second natural hypothesis is that a more formal \textit{distance} measure can better explain the extent of performance degradation for different task-switches. Experiments in Appendix \ref{sec:app-td} explore a ranking based approach as per a range of powerful Large Language Models (LLMs) to obtain some measure of the \textit{distance} between tasks. The empirical results show a weak correlation, which perhaps leads to the conclusion that the performance degradation cannot be simply explained by the specific task-switch but is also a function of the specific model. Future work is required to further understand comprehensively the components that explain the variations in performance degradation for different task-switches and different models..





\section{Conclusions and Future Work}
This work formalizes and investigates the sensitivity of large language models (LLMs) to task-switch scenarios within conversational contexts. We introduce a task-sensitivity as a reference-free metric that can explain a model's behavior to task-switches along with the performance change. By experimenting with various task-switch settings, we observe that even advanced models like GPT-4 can exhibit vulnerabilities to task-switches. Our work additionally lays the foundation for future work on `side-channel' vulnerabilities of LLMs to undesired information leakage/bias from the conversation history. Further work will focus on developing adaptive context management strategies within LLMs to mitigate the risk of task-switch sensitivity. 
\section{Limitations}
Although both GPT-3.5 and GPT-4 show degradation in performance, given the closed nature of OpenAI models, we were not able to perform task sensitivity analysis. We were additionally limited by the maximum token length, hence analysis over extremely long conversations was not feasible. Future work could also look into alignment between humans and the model as a metric which was out of the scope for this paper. 
\section{Ethics and Risks}
All of the datasets used are publicly available. Our implementation utilizes the PyTorch 1.12 framework, an open-source library. We obtained a license from Meta to employ the Llama-7B model via HuggingFace. Additionally, our research is conducted per the licensing agreements of the Mistral-7B, GPT-3.5, and GPT-4 models. We ran our experiments on A100 Nvidia GPU and via OpenAI API. 

Our work may be built upon to identify vulnerabilities of LLMs.  Overall, there are no ethical concerns with this work.

\section{Acknowledgements}

This work was partially supported by Cambridge University 
Press \& Assessment (CUP\&A), a department of
The Chancellor, Masters, and Scholars of the University of Cambridge. This work was partially funded by ELSA – European Lighthouse on Secure and Safe AI funded by the European Union under grant agreement No. 101070617. Views and opinions expressed are however those of the authors only and do not necessarily reflect those of the European Union or European Commission. Neither the European Union nor the European Commission can be held responsible for them. This work was partially funded by the German Federal Ministry of Education and Research (BMBF) under the grant AIgenCY (16KIS2012).

\newpage

\bibliography{custom}
\appendix
\clearpage

\section*{Appendix}

Appendix~\ref{sec:app-dataset-summary} gives more details about the datasets, Appendix~\ref{sec:app-performance} reports the zero-shot absolute performance of all models on all tasks, Appendix~\ref{sec:app-history-length} presents an ablation study on the conversation history length (with multiple seeds), Appendix~\ref{sec:app-prompt} discusses the prompt templates, Appendix~\ref{sec:app-metric} discusses other definitions for task-switch sensitivity, Appendix~\ref{sec:app-correlation} discusses correlations, Appendix~\ref{sec:confuse} tabulates confusion matrices for each model, Appendix~\ref{sec:freerun} investigates task-switch without teacher-forcing the responses in the history, and Appendix~\ref{sec:rand_hist} studies the impact of a randomly generated conversation history. 

\section{Datasets and Metrics Summary} \label{sec:app-dataset-summary}

\begin{table}[htb!]
    \centering
    \small
    \begin{tabular}{lccc}
    \toprule
        Data & \#Train & \#Test & Task  \\ \midrule
       MMLU HA & 26 & 222 & Social MCQ \\
       MMLU AA & 14 & 99 & Math MCQ \\
       RT & 8.53k & 1.07k & Sentiment class \\
       Gigaword  & 3.8M & 1.95k & Summarization \\
       TweetQA & 4.54k & 583 & Social QA\\
       \bottomrule
    \end{tabular}
    \caption{Dataset Summary. QA: Question-Answering. MCQ: Multiple Choice Question}
    \label{tab:data-split}
\end{table}

In Section \ref{sec:results} of the main paper, we present results evaluated on two different datasets: MMLU Abstract Algebra (MMLU AA) multiple choice questions and Rotten Tomatoes (RT) sentiment classification. In Appendix \ref{sec:app-performance}, \ref{sec:app-history-length}, we present results evaluated on all of the datasets covering a range of tasks: MMLU Human Aging (MMLU HA) multiple choice questions, Gigaword for summarization, and TweetQA question-answering. The train-test splits of these datasets are shown in Table ~\ref{tab:data-split}. The train set is randomly sampled to form prompts to produce a conversation history $\mathbf{h}$ of $L$ turns, and the test set is used to evaluate model performance on the $(L+1)$-th turn. The prompt templates used for each dataset are discussed in Appendix~\ref{sec:app-prompt}.

For classification tasks performance is measured using accuracy, whilst for generative tasks it is measured using ROUGE~\citep{lin-2004-rouge} or METEOR~\citep{banarjee2005}.

\section{Absolute Performance} \label{sec:app-performance}

When evaluating the target task with a conversation history, it is useful to compare the performance against a baseline with no conversation history ($\mathbf{h} = \text{Ø}, L = 0$). This is equivalent to evaluating in a zero-shot setting. This section reports the zero-shot performance for all the target task ($T_t$) datasets: MMLU HA in Table ~\ref{tab:mmlu-age-zero_shot}, MMLU AA in Table~\ref{tab:mmluaa-zero_shot}, RT in Table~\ref{tab:rt-zero_shot}, Gigaword in Table ~\ref{tab:gw-zero_shot} and TweetQA in Table~\ref{tab:tq-zero_shot}. Also note that for the classification tasks (MMLU HA, MMLU AA, RT), we also report the number of responses for which we were unable to extract the answer (\# Format Errors), which is further discussed in Appendix \ref{sec:app-prompt}. We evaluated the test set with four LLMs (GPT-3.5, GPT-4, Mistral-7B, Llama-7B), which were all set to temperature 0 for reproducibility.

\begin{table}[htb!]
    \centering
    \small
    \begin{tabular}{l|S[table-format=2.2]S[table-format=2]}
    \toprule
Model & {Accuracy} & {\# Format Errors} \\
\midrule
GPT-3.5 & 66.22 & 18 \\
GPT-4 & 84.68 & 0 \\
Llama-7B & 45.50 & 12 \\
Mistral-7B & 55.41 & 0 \\
       \bottomrule
    \end{tabular}
    \caption{Zero-shot performance on \textbf{MMLU HA}.}
    \label{tab:mmlu-age-zero_shot}
\end{table}

\begin{table}[htb!]
    \centering
    \small
    \begin{tabular}{l|S[table-format=2.2]S[table-format=2]}
    \toprule
Model & {Accuracy} & {\# Format Errors} \\
\midrule
GPT-3.5 & 31.31 & 7 \\
GPT-4 & 58.59 & 0 \\
Llama-7B & 28.28 & 3 \\
Mistral-7B & 21.21 & 0 \\
       \bottomrule
    \end{tabular}
    \caption{Zero-shot performance on \textbf{MMLU AA}.}
    \label{tab:mmluaa-zero_shot}
\end{table}

\begin{table}[htb!]
    \centering
    \small
    \begin{tabular}{l|S[table-format=2.2]S[table-format=1]}
    \toprule
Model & {Accuracy} & {\# Format Errors} \\
\midrule
GPT-3.5 & 89.90 & 0 \\
GPT-4 & 91.80 & 4 \\
Llama-7B & 87.43 & 1 \\
Mistral-7B & 86.68 & 1 \\
       \bottomrule
    \end{tabular}
    \caption{Zero-shot performance on \textbf{RT}.}
    \label{tab:rt-zero_shot}
\end{table}

\begin{table}[htb!]
    \centering
    \small
    \begin{tabular}{l|S[table-format=3.2]S[table-format=1.2]S[table-format=1.2]}
    \toprule
Model & {ROUGE-1} & {ROUGE-2} & {ROUGE-L} \\
\midrule
GPT-3.5 & 17.37 & 4.79 & 14.78 \\
GPT-4 & 15.76 & 4.07 & 13.34 \\
Llama-7B & 11.61 & 3.13 & 9.90 \\
Mistral-7B & 18.60 & 5.19 & 15.84 \\
       \bottomrule
    \end{tabular}
    \caption{Zero-shot performance on \textbf{Gigaword}.}
    \label{tab:gw-zero_shot}
\end{table}

\begin{table}[htb!]
    \centering
    \small
    \begin{tabular}{l|S[table-format=2.2]S[table-format=2.2]S[table-format=2.2]}
    \toprule
Model & {ROUGE-1} & {ROUGE-L} & {METEOR} \\
\midrule
GPT-3.5 & 30.66 & 30.39 & 44.18 \\
GPT-4 & 28.03 & 27.68 & 43.41 \\
Llama-7B & 17.91 & 17.67 & 33.84 \\
Mistral-7B & 25.35 & 25.01 & 40.71 \\
       \bottomrule
    \end{tabular}
    \caption{Zero-shot performance on \textbf{TweetQA}.}
    \label{tab:tq-zero_shot}
\end{table}

\newpage
\section{Conversation History Length Ablation} \label{sec:app-history-length}

This section presents an ablation study on the performance change after a task-switch for varying conversation history lengths. For each dataset in Table~\ref{tab:data-split} we select four datasets (including itself), from which we use the training set as conversation history. The details of the prompt structure are presented in Appendix~\ref{sec:app-prompt}. 

\subsection{Task-switch Performance Change} \label{sec:app-perf-change}

\begin{figure*}[h]
    \centering
    \includegraphics[width=\textwidth]{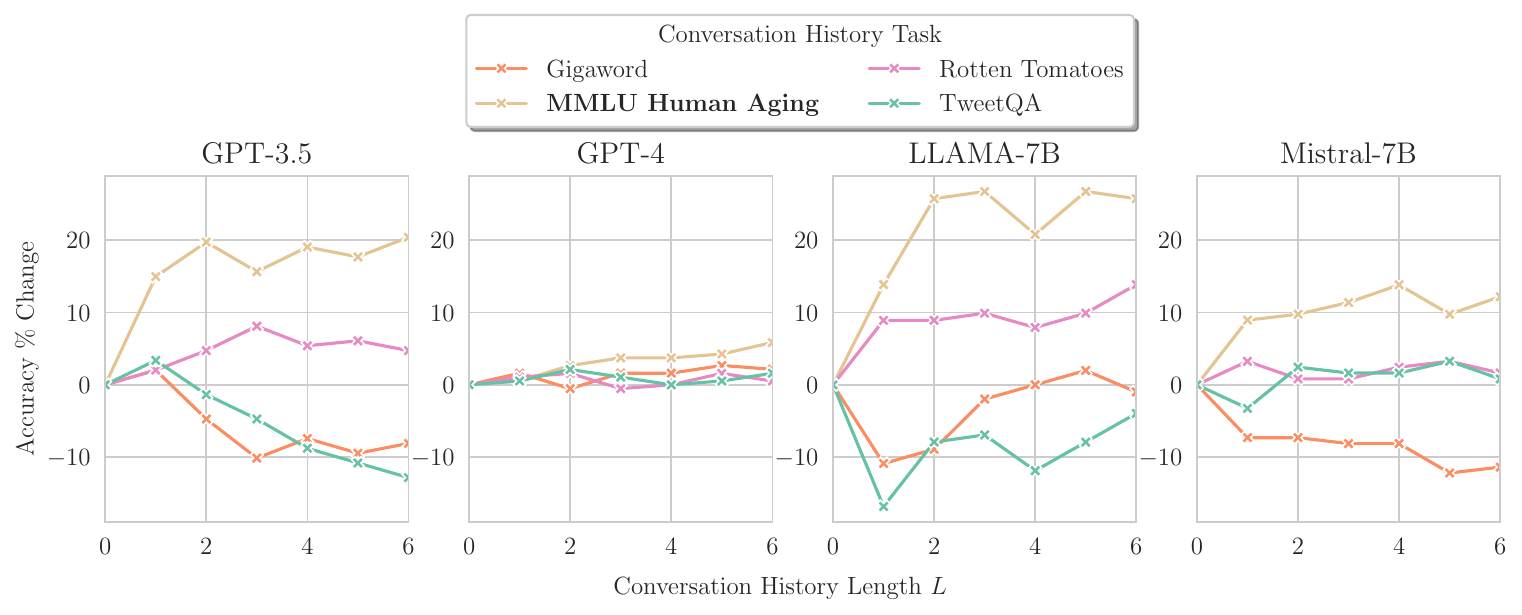}
    \caption{Target Task: MMLU HA. Percentage \% change in accuracy relative to zero-shot performance (no conversation history) for increasing conversation history length $L$ and various models.}
    \label{fig:mmlu-age-acc}
\end{figure*}

\begin{figure*}[h]
    \centering
    \includegraphics[width=\textwidth]{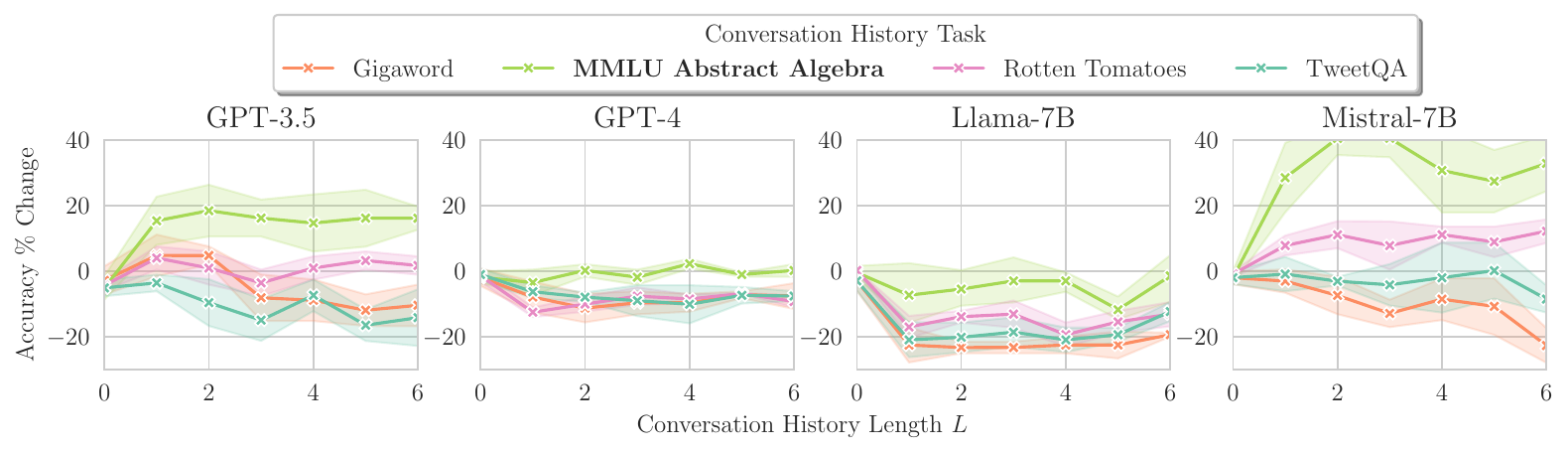}
    \caption{Target Task: MMLU AA. Percentage \% change in accuracy relative to zero-shot performance (no conversation history) for increasing conversation history length $L$ and various models. }
    \label{fig:mmluaa-acc-app}
\end{figure*}

\begin{figure*}[h]
    \centering
    \includegraphics[width=\textwidth]{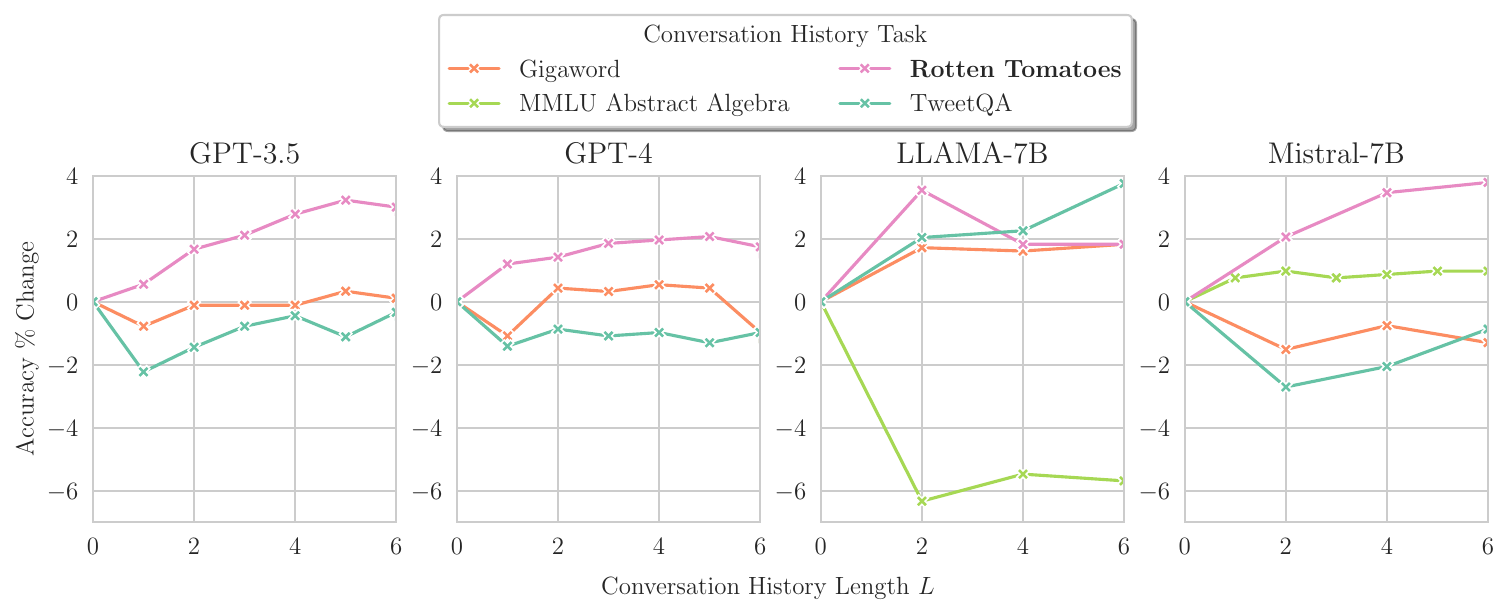}
    \caption{Target Task: RT. Percentage \% change in accuracy relative to zero-shot performance (no conversation history) for increasing conversation history length $L$ and various models. }
    \label{fig:rt-acc}
\end{figure*}

\begin{figure*}[h]
\centering
\begin{subfigure}{\linewidth}
    \caption{ROUGE-1}
    \includegraphics[width=\linewidth]{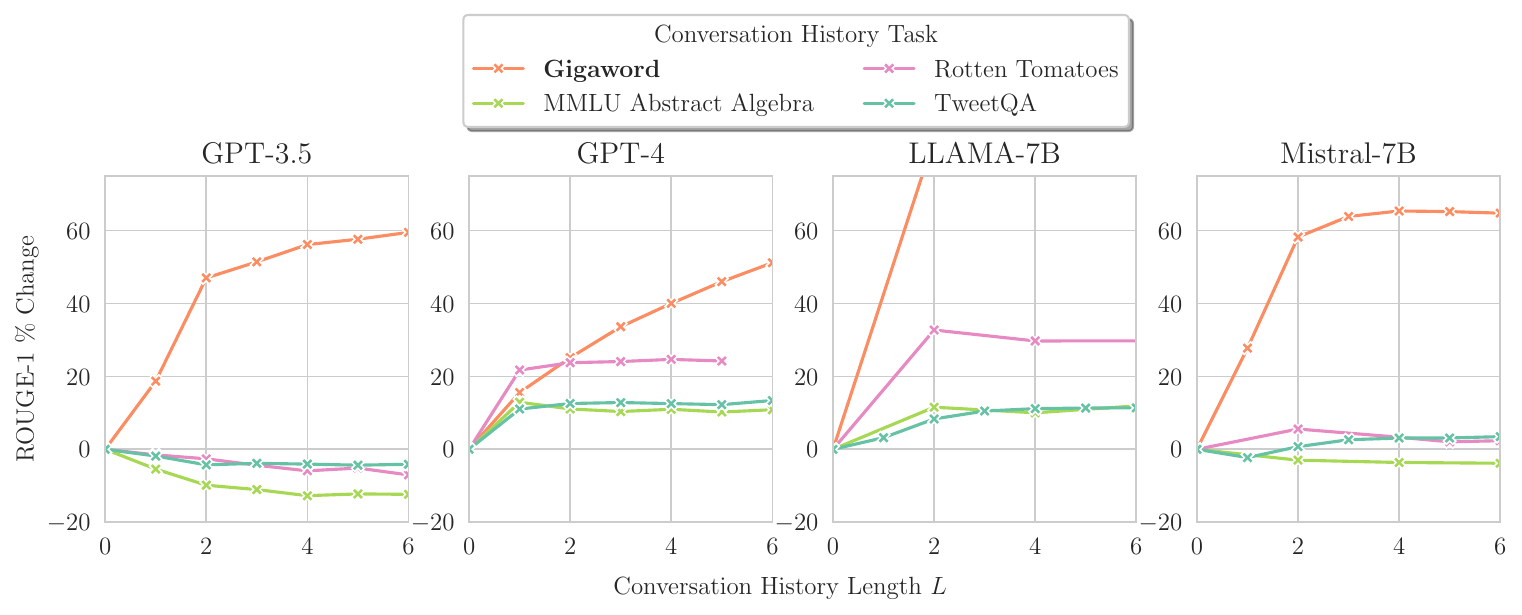}
    \label{fig:gw-rouge1}
\end{subfigure}
\begin{subfigure}{\linewidth}
    \caption{ROUGE-2}
    \includegraphics[width=\linewidth]{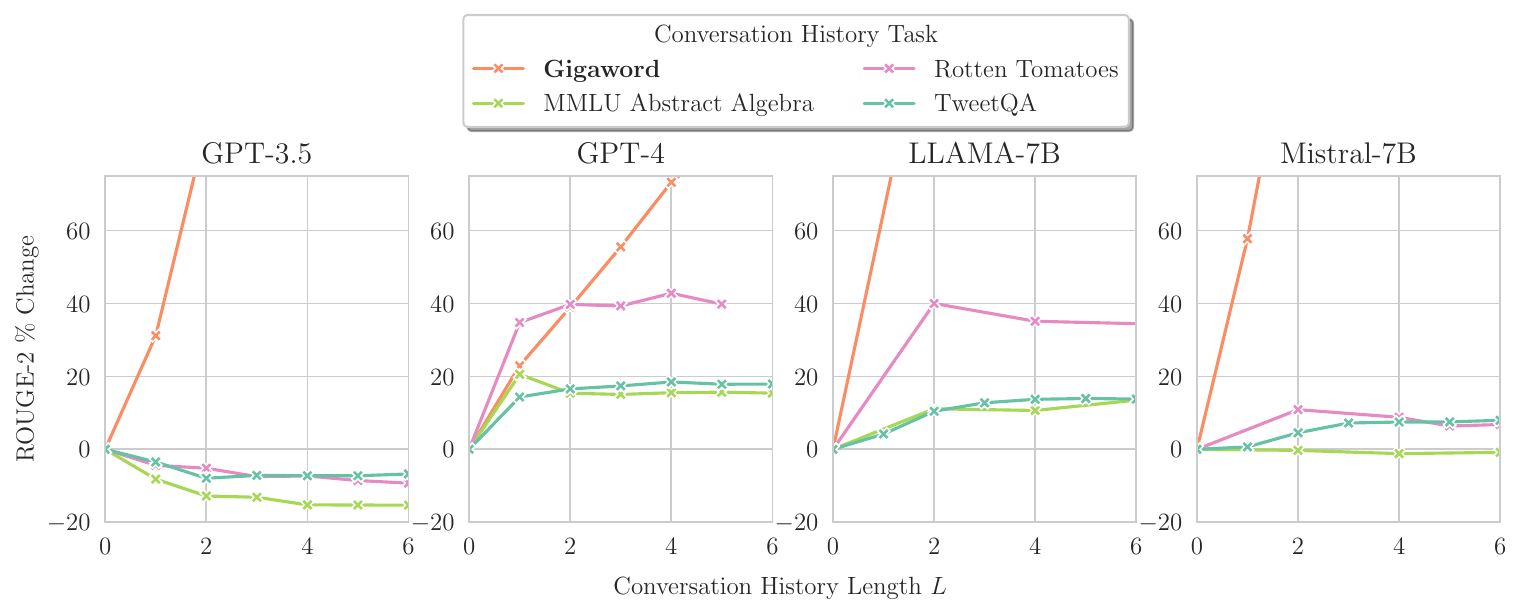}
    \label{fig:gw-rouge2}
\end{subfigure}
\begin{subfigure}{\linewidth}
    \caption{ROUGE-L}
    \includegraphics[width=\linewidth]{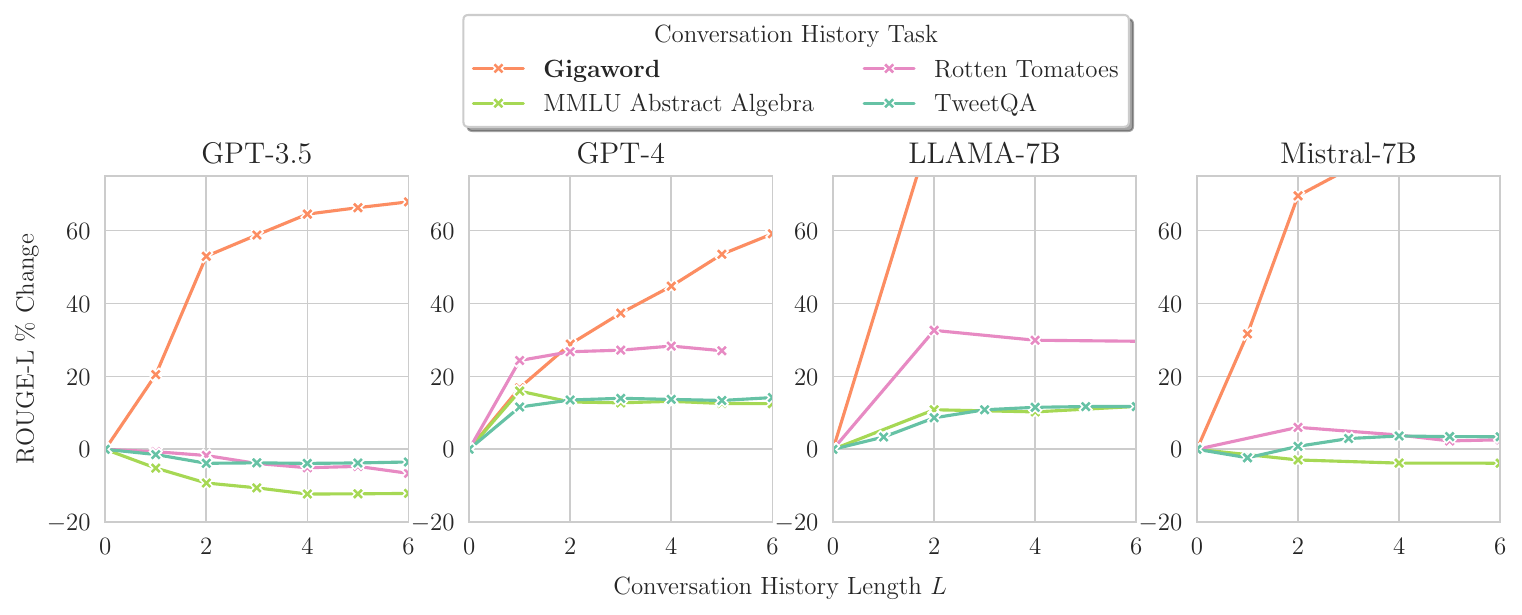}
    \label{fig:gw-rougel}
\end{subfigure}
\caption{Target Task: Gigaword. Percentage \% change in accuracy relative to zero-shot performance (no conversation history) for increasing conversation history length $L$ and various models. Note that we focus on the effect of task-switching by clipping the y-axes at +75\%.}
\label{fig:gw-acc}
\end{figure*}

\begin{figure*}[h]
\centering
\begin{subfigure}{\linewidth}
    \caption{ROUGE-1}
    \includegraphics[width=\linewidth]{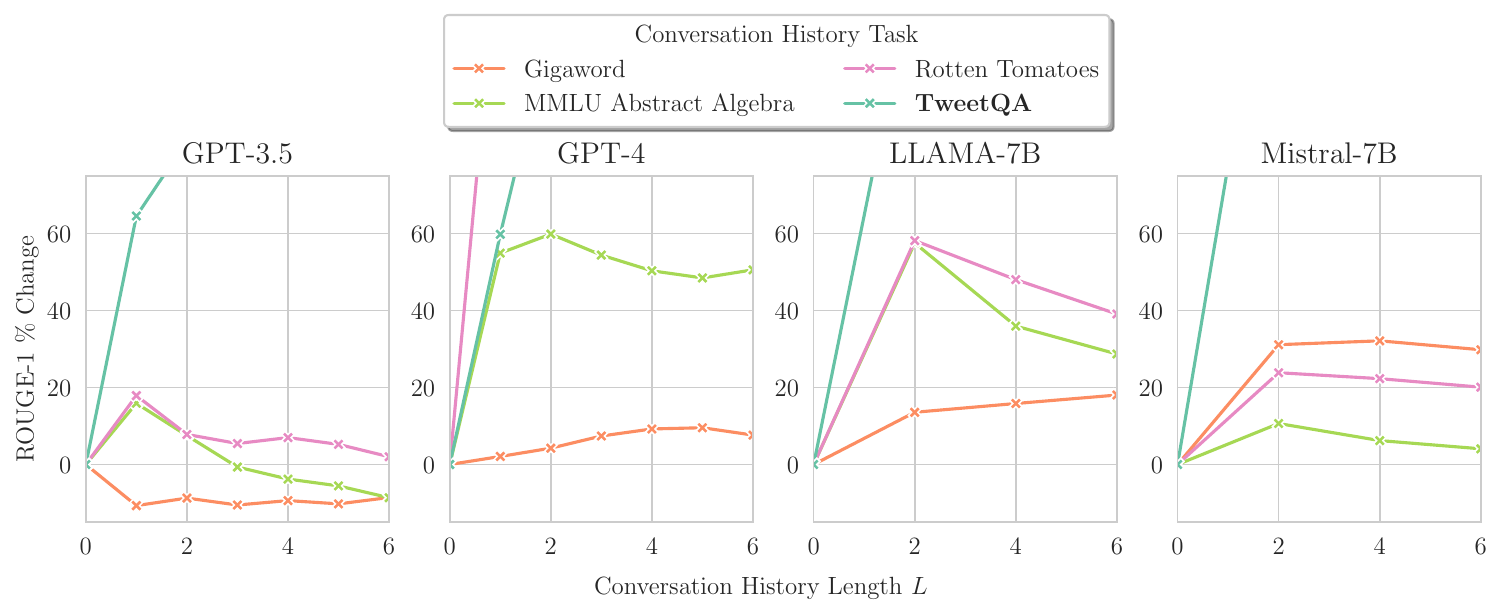}
    \label{fig:tq-rouge1}
\end{subfigure}
\begin{subfigure}{\linewidth}
    \caption{ROUGE-L}
    \includegraphics[width=\linewidth]{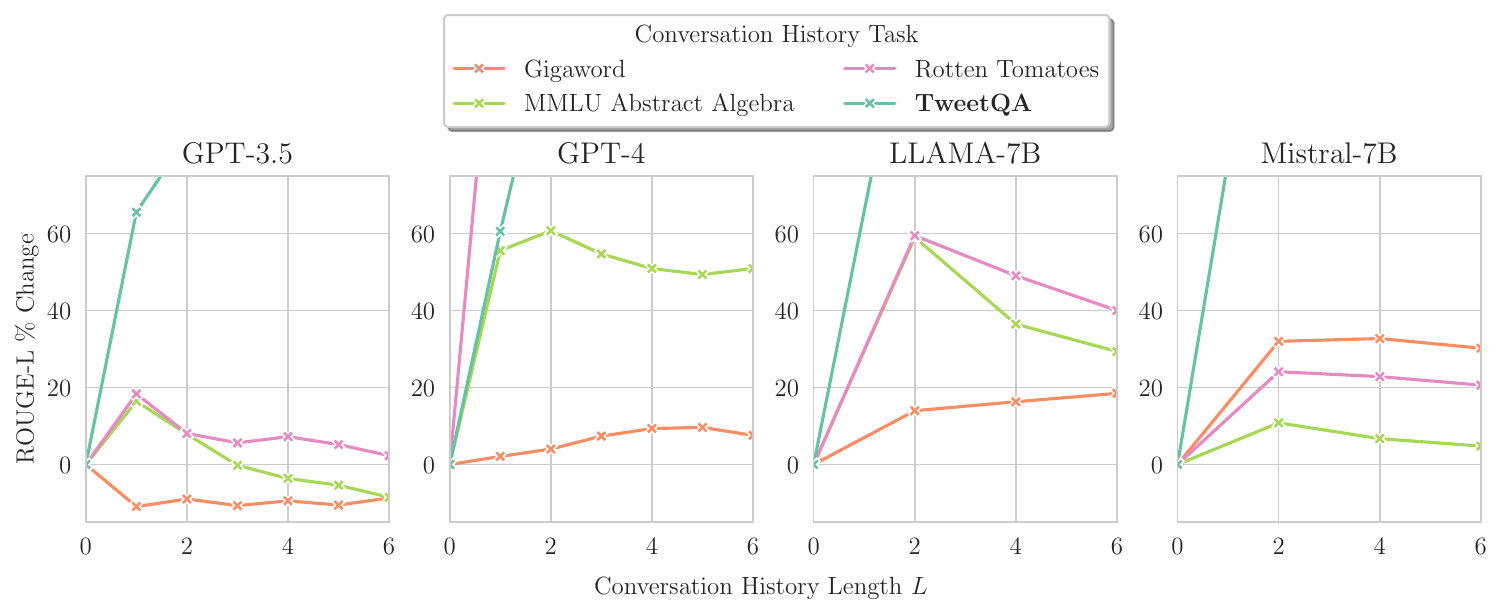}
    \label{fig:tq-rougel}
\end{subfigure}
\begin{subfigure}{\linewidth}
    \caption{METEOR}
    \includegraphics[width=\linewidth]{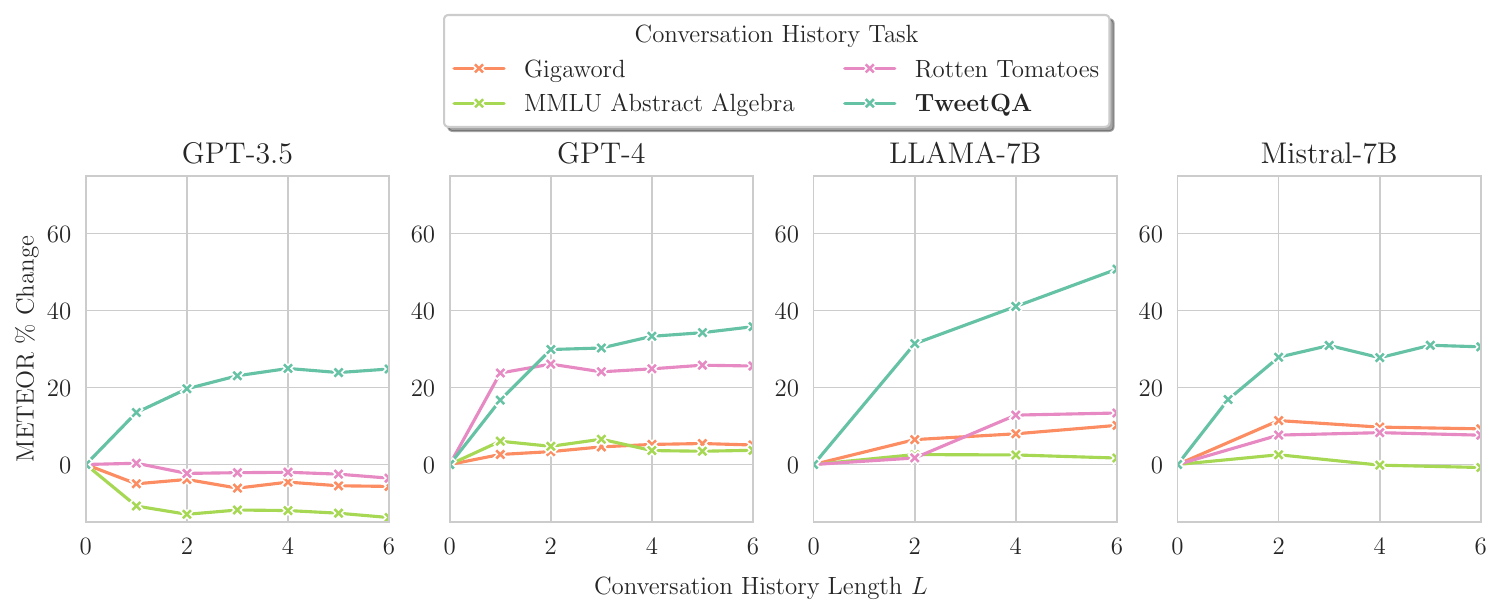}
    \label{fig:tq-meteor}
\end{subfigure}
\caption{Target Task: TweetQA. Percentage \% change in accuracy relative to zero-shot performance (no conversation history) for increasing conversation history length $L$ and various models. Note that we focus on the effect of task-switching by clipping the y-axes at +75\%.}
\label{fig:tq-acc}
\end{figure*}

We compare the percentage change in metrics relative to zero-shot performance ($\mathbf{h}=\text{Ø}$, i.e. no conversation history) as a function of conversation history length $L$ and for different LLMs. Results are plot in Figures \ref{fig:mmlu-age-acc}, \ref{fig:mmluaa-acc-app}, \ref{fig:rt-acc}, \ref{fig:gw-acc}, \ref{fig:tq-acc} for MMLU HA, MMLU AA, RT, Gigaword and TweetQA respectively. When there is \emph{not} a task switch, we would expect a performance increase (assuming the training examples are representative of
the test set). As per our discussion in Section~\ref{sec:results}, we observe that different models degrade on different task-switches and this is not limited by the model size.

\subsection{Format Failure Rate} \label{sec:app-format-fail}

Typically, classification tasks (MMLU HA, MMLU AA, RT) are evaluated using logits, however we use a generative approach for consistency: we are evaluating the model in a conversational setting, and we do not have access to the logits exactly. Thus, we must post-process the model output to determine the class. In this, we try to give the LLM the benefit of the doubt and do our best to extract the class. For example, although the prompt requests the model to output within answer tags like \texttt{"<Answer> positive </Answer>"}, we also accept \texttt{"positive"}, but we do not accept \texttt{"positive/negative"}. Due to the imperfect nature of this setup, either we may not detect the correct format, or the model generates erroneous text. 

\begin{figure*}[h]
    \centering
    \includegraphics[width=\textwidth]{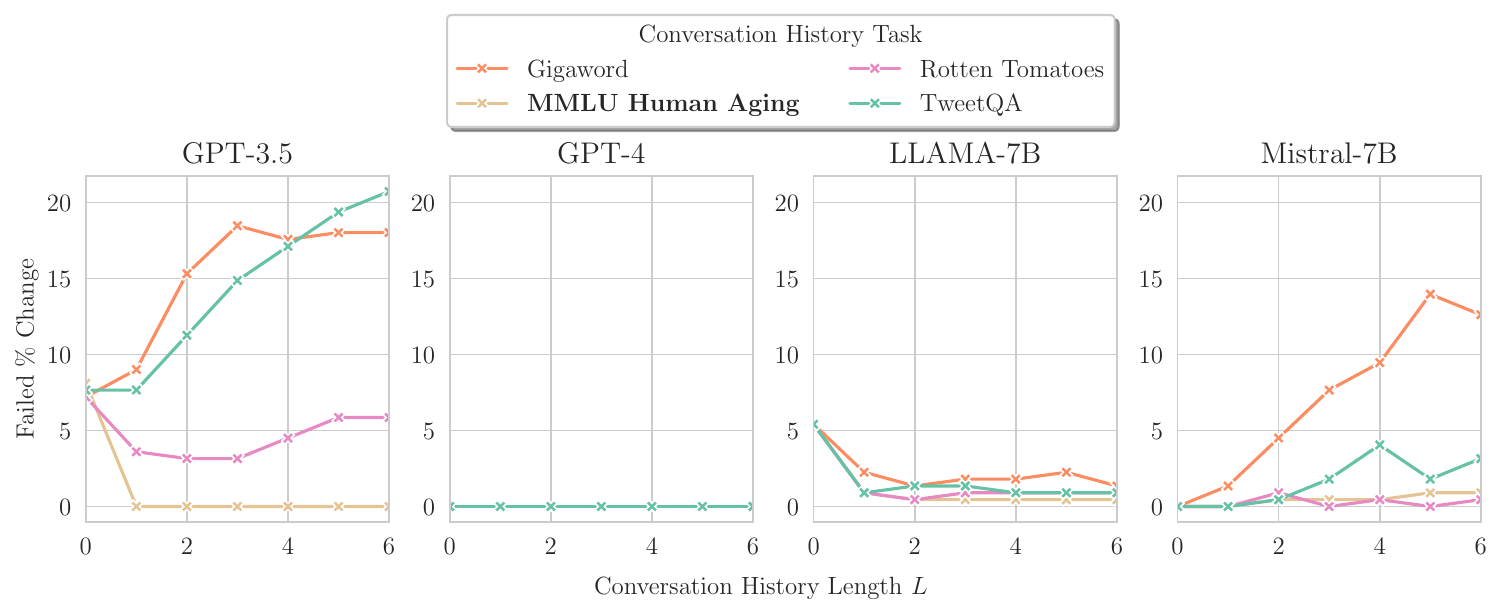}
    \caption{Target Task: MMLU Human Aging. Percentage \% of examples where format failed.}
    \label{fig:mmlu-age-fail}
\end{figure*}

\begin{figure*}[h]
    \centering
    \includegraphics[width=\textwidth]{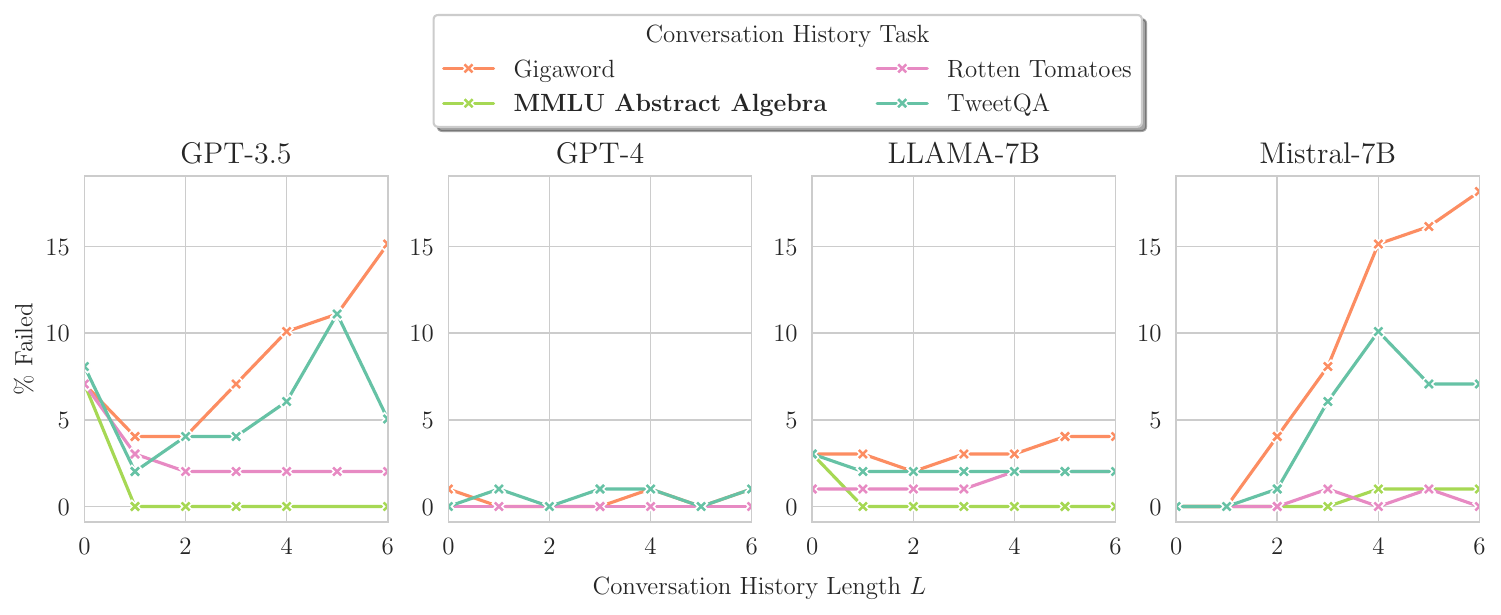}
    \caption{Target Task: MMLU Abstract Algebra. Percentage \% of examples where format failed.}
    \label{fig:mmlu-aa-fail}
\end{figure*}

\begin{figure*}[h]
    \centering
    \includegraphics[width=\textwidth]{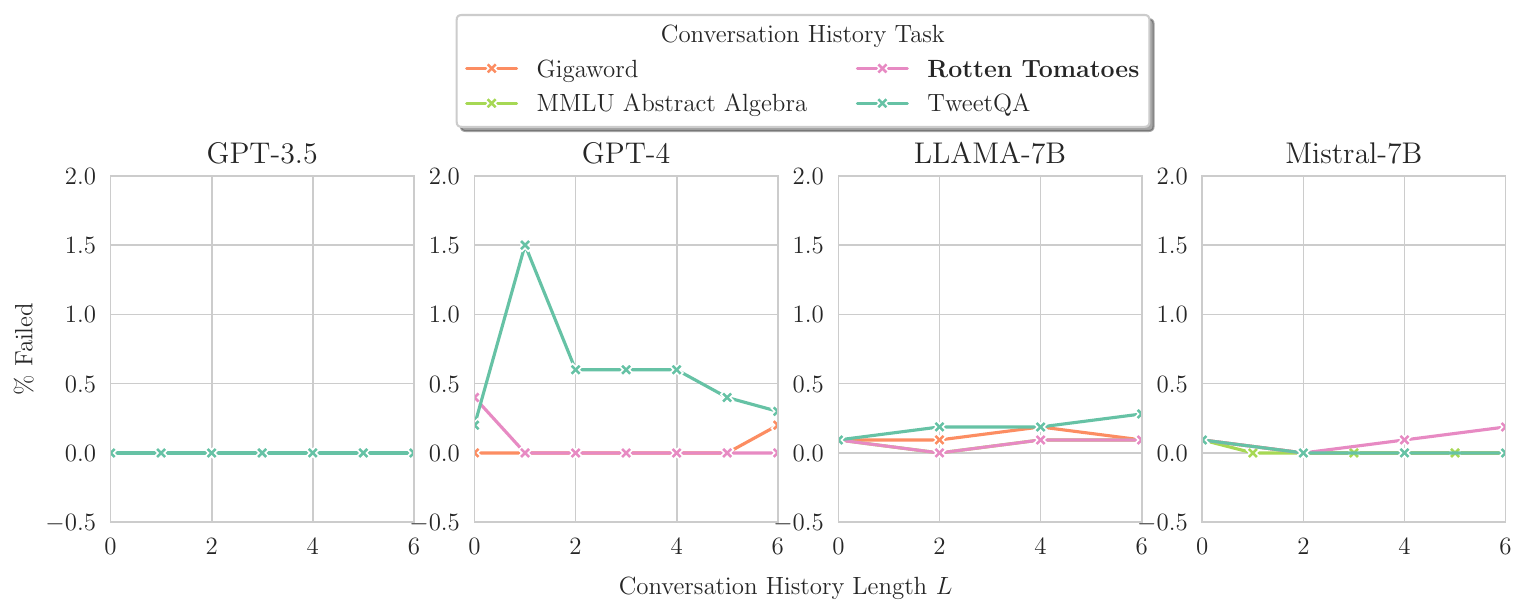}
    \caption{Target Task: Rotten Tomatoes. Percentage \% of examples where format failed.}
    \label{fig:rt-fail}
\end{figure*}

Importantly, models may become more susceptible to these errors when performing a task-switch, causing performance degradation. We capture this by reporting the percentage \% change in the number of examples that the model failed on (relative to zero-shot) as the context history length increases. These are plot in Figures \ref{fig:mmlu-age-fail}, \ref{fig:mmlu-aa-fail}, \ref{fig:rt-fail} for MMLU HA, MMLU AA and RT respectively. Figures \ref{fig:mmlu-age-fail} and \ref{fig:mmlu-aa-fail} show that GPT-3.5 and Mistral-7B are susceptible to format errors in task-switches when evaluating on multiple choice questions, whereas Figure \ref{fig:rt-fail} shows that GPT-4 and Llama-7B are more susceptible in sentiment classification.

\section{Prompt Template} \label{sec:app-prompt}

In each conversation turn, the user prompts the model $u_k$. The prompts are shown in Table~\ref{tab:prompt_table}. We chose these prompts after careful research and experimentation. We began with popular templates and refined them for our purpose.


Additionally, since we do not have access to the logits for all models, we take a generative approach to the classification tasks (MMLU HA, MMLU AA, RT). Since the model may fail to output an answer in the desired format, we post process the text to extract the answer (which we count as a positive result it matches the reference). We report and discuss the effect of format failures further in \ref{sec:app-format-fail}. Furthermore, we note that the standard evaluation method used in the Open-LLM leaderboard code (available on \href{https://github.com/declare-lab/instruct-eval/blob/c85b7532149eacf3646734cfbb06af1de2d27765/mmlu.py#L155}{GitHub}) is to see if the response starts with \texttt{A,B,C} or \texttt{D}\cite{eval-harness}. We modified the prompt to ensure a more consistent output format (across the different models) resulting in fewer mistakes made. 

For the classification tasks, we structure the prompt such that we request the model to output their final answer within answer tags. We note that giving an example of how to use the answer tags always helped, however, this can bias the model towards a particular answer. Instead, we found for MMLU to just leave the answer tags empty, whereas for RT to have the all the sentiment classes inside the tags (see Table \ref{tab:prompt_table} for further details).

\clearpage

\begin{table*}[htb!]
    \centering
    \small
    \begin{tabular}{l|p{10cm}}
    \toprule
    \textbf{MMLU} \{Topic\}& 
        \texttt{You have a multiple choice question on \{Topic\}. Only one of the options is correct: A, B, C, or D. Give your answer in the following format with the tags provided: <Answer> </Answer>. Please read the following question and options and answer the question
        \newline
        Question: \{Question\} \newline
        (A) \{A\} \newline
        (B) \{B\} \newline
        (C) \{C\} \newline
        (D) \{D\}
        } \\
    \midrule
    \textbf{Rotten Tomatoes} & 
        \texttt{Can you choose only one sentiment [`negative', `positive'] for this review.\newline
        review: \{Review\} \newline
        Return only the sentiment label without any other text. Make sure to follow the format otherwise your answer will be disqualified: 
        \newline <Answer> positive / negative </Answer>.
        \newline Do not output neutral.
        } \\
    \midrule
    \textbf{Gigaword} &
        \texttt{Please summarize the following article.\newline
        \{Article\}} \\
    \midrule
    \textbf{TweetQA} &
        \texttt{Read the given tweet and answer the corresponding question.\newline
        tweet: \{Tweet\}\newline
        question: \{Question\}} \\
    \bottomrule
    \end{tabular}
    \caption{Prompt templates for each dataset. Note that the MMLU \{Topic\} can be either \texttt{Human Aging} or \texttt{Abstract Algebra}. Other \{words\} enclosed in curly braces are replaced by the corresponding field in the datasets.}
    \label{tab:prompt_table}
\end{table*}

\begin{figure*}[h]
\centering
\begin{subfigure}{\linewidth}
    \caption{Zero-shot Sensitivity}
    \includegraphics[width=\linewidth]{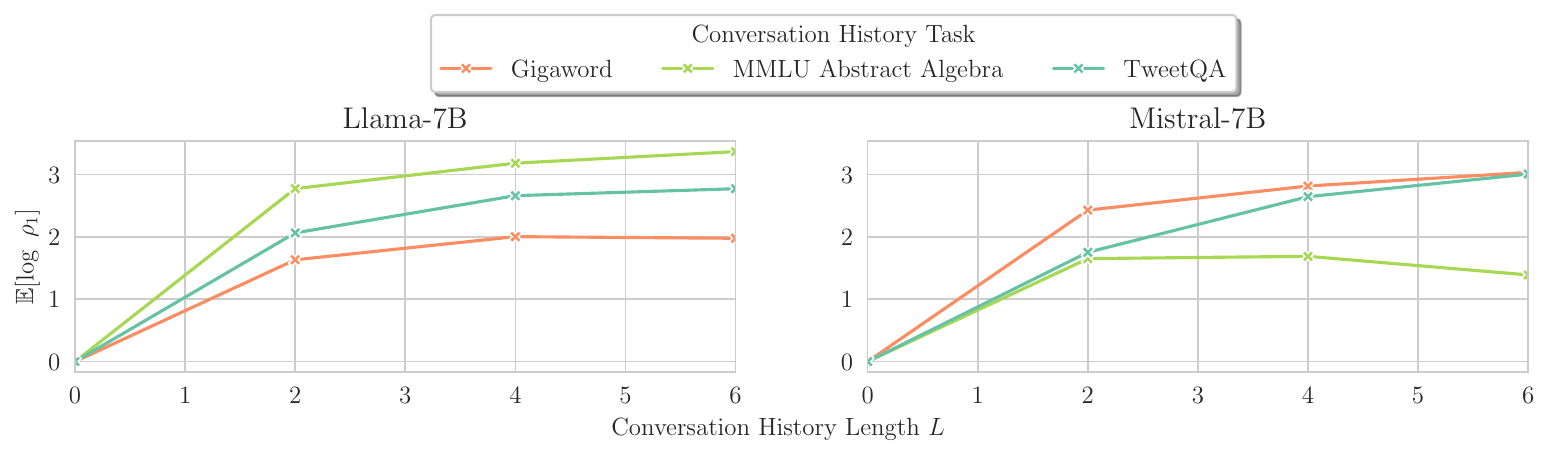}
    \label{fig:sense-zeroshot}
\end{subfigure}
\begin{subfigure}{\linewidth}
    \caption{Confidence Sensitivity}
    \includegraphics[width=\linewidth]{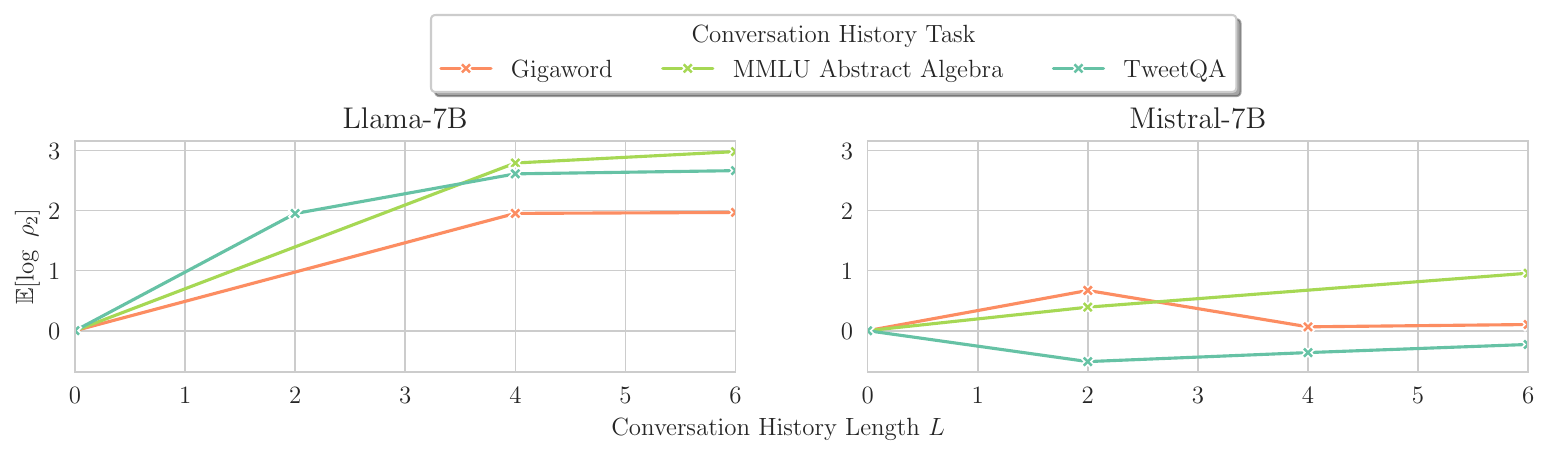}
    \label{fig:sense-conf}
\end{subfigure}
\begin{subfigure}{\linewidth}
    \caption{Loss Sensitivity}
    \includegraphics[width=\linewidth]{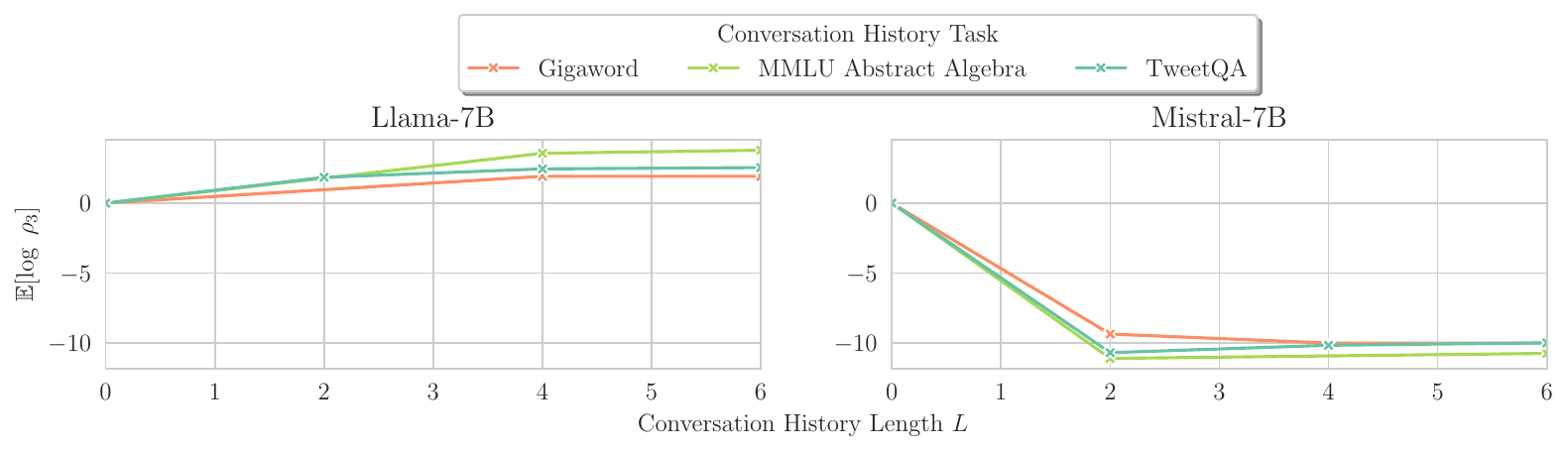}
    \label{fig:sense-loss}
\end{subfigure}
\caption{Empirical investigation of various sensitivity metrics on the target task Rotten Tomatoes as a function of the conversation history length $L$ for Llama-7b and Mistral-7b. Note that we omit the line for the in-context dataset as this is not relevant to the investigation.}
\label{fig:rt-sensitivity}
\end{figure*}

\clearpage
\section{Task-Switch Sensitivity Metrics} \label{sec:app-metric}

In Section \ref{sec:method}, we introduced and formalized evaluation of a model's sensitivity to task-switch, namely the task sensitivity $\tau$. This metric aims to capture the vulnerability of a model prompt to its chat history after a task-switch. Formally, it compares the zero-shot prediction $r^* | u, \mathbf{h} = \text{Ø}$ to the probability of the model outputting the same zero-shot response after a task switch $P(r^* | u, \mathbf{h} \ne \text{Ø})$. In this section, we compare the theoretical and empirical implications of different task switch sensitivity metrics. 

Formally, given a conversation history $\mathbf{h}$ of length $L$ and the next user prompt $u$, the probability of a model's response $r_{L+1}$ is given by $P_\theta (r_{L+1} \ | \ u, \mathbf{h})$. We consider the probability of three possible responses:

\begin{enumerate}
    \item $r^*$: zero-shot response
    \item $r_{L+1}$: model's actual response
    \item $\hat{r}_{L+1}$: reference response
\end{enumerate}

We posit that after a task-switch, a robust model's likelihood of the zero-shot response remains high. Naturally, this gives us the formulation for the aforementioned sensitivity metric
\begin{equation}
    \rho_1 = \frac{P_\theta(r^* | u)}{P_\theta(r^* | u, \text{\textbf{h}})},
\end{equation}
which we call \emph{zero-shot sensitivity}. 

Additionally, after a task-switch, we posit that a robust model's likelihood of the actual response should be similar to that of the zero-shot response, because the irrelevant history should be largely ignored. This gives us
\begin{equation}
    \rho_2 = \frac{P_\theta(r^* | u)}{P_\theta(r_{L+1} | u, \text{\textbf{h}})},
\end{equation}
which we call the \emph{confidence sensitivity}.

Lastly, we posit that if a model is well aligned to a task, then both the zero-shot and model's actual response should be close to the reference response:
\begin{equation}
    \rho_3 = \frac{P_\theta(\hat{r}_{L+1} | u)}{P_\theta(\hat{r}_{L+1} | u, \text{\textbf{h}})},
\end{equation}
where each probability is essentially a measure of the loss, hence we label this as the \emph{loss sensitivity}.

The above are sensitivity per example, which we can use to estimate the task-switch sensitivity $\tau_i = \mathbb{E}[\log \ \rho_i]$ as per Equation \ref{eqn:tau}, where the expectation is calculated over the examples and histories (for a given length $L$). 
We evaluate these metrics on the target task RT (rotten tomatoes) as shown in Figure~\ref{fig:rt-sensitivity}. 
Figure~\ref{fig:sense-zeroshot} shows that the zero-shot sensitivity metric trends upwards for both models. This is expected for a model which does not handle task-switch well as the probability of the output with an increased conversation length decreases in comparison to the zero-shot probability.
For the confidence sensitivity in Figure~\ref{fig:sense-conf}, we observe that Mistral-7B behaves as we expect, whereas Llama-7B becomes less confident in its output compared to having no conversation history. For the loss sensitivity metric in Figure~\ref{fig:sense-loss}, we observe that Llama behaves as we expect as the sensitivity remains relatively flat: as the conversation history increases, there is no significant change in the probability of outputting the reference. However, for Mistral-7B, the probability falls immediately and plateaus showing that the model was giving a very low probability mass to the reference with no conversation history.  
Intuitively, it is clear that both models agree in their trends only for the zero-shot sensitivity $\tau_1$ in Figure~\ref{fig:sense-zeroshot}, hence in the main paper, we report zero-shot sensitivity as the task-switch sensitivity.  Note finally that a further major advantage of the zero-shot sensitivity is that it is reference-free and so can be used in practice to predict the vulnerability of deployed models to different task-swithches when there is no access to the correct, reference response.

\newpage
\section{Correlations Models, Datasets and Performance}
\label{sec:app-correlation}

We rank model performance against various metrics to see if there is any correlation that may help explain model performance more generally.

\subsection{Task Conversation History Length} \label{sec:app-tchl}

\begin{table}[!htb]
    \centering
    \small
    \begin{tabular}{l|c|c|c}
    \toprule
    CH Task & Length & Llama-7B & Mistral-7B \\
    \midrule
    Gigaword & 75 & -21.35 & -15.94 \\
    TweetQA & 93 & -15.10 & -4.35 \\
    RT & 108 & -13.02 & 10.87 \\
    MMLU AA & 143 & -1.79 & 37.68 \\
    \bottomrule
    \end{tabular}
    \caption{Target Task: \textbf{MMLU AA}. Average length of response (length) in Conversation History (CH) compared to the performance degradation of models in the task-switch.}
    \label{tab:mmluaa-task-length-vs-performance}
\end{table}

\begin{table}[!htb]
    \centering
    \small
    \begin{tabular}{l|c|c|c}
    \toprule
    CH Task & Length & Llama-7B & Mistral-7B \\
    \midrule
    Gigaword & 76 & 1.98 & -0.72 \\
    TweetQA & 93 & 2.70 & -1.28 \\
    RT & 108 & 2.38 & 2.83 \\
    MMLU AA & 143 & -5.42 & 1.19 \\
    \bottomrule
    \end{tabular}
    \caption{Target Task: \textbf{RT}. Average length of response (length) in Conversation History (CH) compared to the performance degradation of models in the task-switch.}
    \label{tab:rt-task-length-vs-performance}
\end{table}

In Table \ref{tab:mmluaa-task-length-vs-performance} and Table \ref{tab:rt-task-length-vs-performance} we compare the model performance against the mean conversation history task, $T_h$ length, which is the average number of tokens per response in a turn in the conversation history. The model performance is taken for three different seeds with conversation history lengths $L \in \{3,6\}$. There is no observed correlation between the average length of responses for a task in a conversation and the performance degradation observed for different models.

\subsection{Task Distance} \label{sec:app-td}

In this section we aim to assess the hypothesis that the `distance' between tasks can explain the extent of performance degradation in different task-switches, from the conversation history task, $T_h$ to the target task, $T_t$. Measuring distance between tasks is a multi-faceted and complex metric. Given the lack of formal task distance measures, we instead use a consensus ranking approach, where multiple powerful Large Language Models (LLMs) are required to rank the different tasks on how similar they are. For the target task \textit{RT}, we queried four of the largest and most powerful models to rank the closest tasks, based on the description of each task. We consider the following LLMs: ChatGPT; Gemini Ultra~\citep{geminiteam2024gemini}, Claude 3 Sonnet from Anthropic; and Perplexity AI. The rankings by the LLMs are given in Table \ref{tab:llm-rank} relative to RT. We then select an overall ranking with the greatest consensus - in this case three of the four LLMs agree perfectly in the ranking. This gives a consensus vote of ranks (relative to RT): RT (1); MMLU AA (3); TweetQA (2); and Gigaword (3). The equivalent ranks are given in Table \ref{tab:llm-rank-mmlu} with MMLU AA as the reference task. In this case, three of the four models perfectly agree in their rankings.

\begin{table}[htb!]
    \centering
    \scriptsize
    \begin{tabular}{l|cccc}
    \toprule
       Dataset  &  \tiny{ChatGPT} & \tiny{Gemini} & \tiny{Claude} & \tiny{Perplexity} \\ \midrule
       RT & 1 & 1 & 1 & 1 \\
        MMLU AA & 4 & 4 & 4 & 4 \\
        TweetQA & 2 & 3 & 2 & 2 \\
        Gigaword & 3 & 2 & 3 & 3 \\
        \bottomrule
    \end{tabular}
    \caption{Rank given by LLM for different datasets on how similar they are to the target task RT.}
    \label{tab:llm-rank}
\end{table}

\begin{table}[htb!]
    \centering
    \small
    \begin{tabular}{l|cccc}
    \toprule
       Dataset  &  \tiny{ChatGPT} & \tiny{Gemini} & \tiny{Claude} & \tiny{Perplexity}\\ \midrule
        RT & 4 & 4 & 4 & 4 \\
        MMLU AA & 1 & 1 & 1 & 1 \\
        TweetQA & 2 & 3 & 2 & 2 \\
        Gigaword & 3 & 2 & 3 & 3 \\
        \bottomrule
    \end{tabular}
    \caption{Rank given by LLM for different datasets on how similar they are to the target task MMLU AA.}
    \label{tab:llm-rank-mmlu}
\end{table}

The following tables compare the rank of the dataset distance against the mean model performance. The model performance is the \%-percentage accuracy change relative to zero-shot, and the mean is taken over three seeds and over conversation history lengths $L \in \{3,6\}$.

\begin{table}[!htb]
    \small
    \centering
    \begin{tabular}{l|c|c|c}
    \toprule
    CH-Task & Rank & Llama-7B & Mistral-7B \\
    \midrule
    RT       & 1 & 2.38  & 2.83 \\
    TweetQA  & 2 & 2.70  & -1.28 \\
    Gigaword & 3 & 1.98  & -0.72 \\
    MMLU AA  & 4 & -5.42 & 1.19 \\
    \bottomrule
    \end{tabular}
    \caption{Target Task, $T_t$: \textbf{RT}. Performance degradation (with different conversation history tasks) compared to the task rank, measuring similarity to $T_t$.}
    \label{tab:rt-task-distance-vs-performance}
\end{table}

\begin{table}[!htb]
    \small
    \centering
    \begin{tabular}{l|c|c|c}
    \toprule
    CH-Task & Rank & Llama-7B & Mistral-7B \\
    \midrule
    MMLU AA  & 1 & -1.79 & 37.68 \\
    TweetQA  & 2 & -15.10 & -4.35 \\
    Gigaword & 3 & -21.35 & -15.94 \\
    RT       & 4  & -13.02 & 10.87 \\
    \bottomrule
    \end{tabular}
    \caption{Target Task: \textbf{MMLU AA}. Performance degradation (with different conversation history tasks) compared to the task rank, measuring similarity to $T_t$.}
    \label{tab:mmluaa-rank-vs-performance}
\end{table}

Overall, there appears to be only a weak correlation in some settings between the task distance and the performance degradation. This suggests that performance degradation is not only a function of the task distance, but is also an attribute of the specific model. Further analysis would be required to understand the aspects of specific models for certain task-switches that influence the level of performance degradation. 

\section{Performance Confusion Matrix}
\label{sec:confuse}

In this section, we summarize the performance change for every pairing of task-switches from conversation history task ($T_h$) to target task ($T_t$). We present the results here for a conversation length of $L=6$ for each model separately. Tables \ref{tab:gpt3.5-confuse}, \ref{tab:gpt4-confuse}, \ref{tab:llama-confuse}, \ref{tab:mistral-confuse} report the results for models GPT-3.5, GPT-4, Llama-7B, Mistral-7B respectively. Each \emph{row} is the performance change in the Target Task $T_t$. Please note that the metric for the tasks are: accuracy for MMLU AA, RT, MMLU HA, METEOR for TweetQA, and RougeL for Gigaword.

\begin{table}[!htb]
    \centering
    \small
    \begin{tabular}{>{\tiny}l|c|c|c|c|c}
    \toprule
     & \multicolumn{5}{|c}{Conversation History Task} \\
     Target Task & AA & RT & TQ & GW & HA \\
     \midrule
      {MMLU AA}  & 	19.35 &	6.45 &	6.45 &	-3.13 &	*\\
     {RT}  & 	-0.22 &	3.00 &	-0.33 &	0.11 & *\\
     Tweet QA  & 	-13.78 &	-3.55 &	24.81 &	-5.69 & *\\
     Gigaword  & 	-12.10 &	-6.59 &	-3.48 &	67.85 & *\\
     MMLU HA  & *	 & 4.73 &	-12.84 &	-8.11 &	20.41 \\ \bottomrule
    \end{tabular}
    \caption{Model: \textbf{GPT-3.5}. Percentage \% change in model performance.}
    \label{tab:gpt3.5-confuse}
\end{table}

\begin{table}[!htb]
    \centering
    \small
    \begin{tabular}{>{\tiny}l|c|c|c|c|c}
    \toprule
     & \multicolumn{5}{|c}{Conversation History Task} \\
     Target Task & AA & RT & TQ & GW & HA \\
     \midrule
    MMLU AA  & 	8.62 &	-13.11 &	-3.39 &	0.00 & *\\
    RT  & 	0.76 &	1.74 &	-0.98 &	-0.98 & *\\
     Tweet QA  & 	3.69 &	25.58 &	35.80 &	5.06 &	* \\
     Gigaword  & 	12.52 &	* & 14.18 &	59.07 & *\\
     MMLU HA  & 	* & 0.53 &	1.59 &	2.14 &	5.85 \\ \bottomrule
    \end{tabular}
    \caption{Model: \textbf{GPT-4}. Percentage \% change in model performance.}
    \label{tab:gpt4-confuse}
\end{table}

\begin{table}[!htb]
    \centering
    \small
    \begin{tabular}{>{\tiny}l|c|c|c|c|c}
    \toprule
     & \multicolumn{5}{|c}{Conversation History Task} \\
     Target Task & AA & RT & TQ & GW & HA \\
     \midrule
     MMLU AA  & 	3.57 &	-15.63 &	0.00 &	-10.71 & *\\
     RT  & 	-5.69 &	1.82 &	3.76 &	1.82 & *\\
     Tweet QA  & 	1.68 &	13.37 &	50.74 &	10.17 & *\\	
     Gigaword  & 	11.76 &	 * & 11.73 &	158.79 & *\\
     MMLU HA  & 	* & 13.86 &	-3.96 &	-0.99 &	25.74 \\ 
     \bottomrule
    \end{tabular}
    \caption{Model: \textbf{Llama-7B}. Percentage \% change in model performance.}
    \label{tab:llama-confuse}
\end{table}

\begin{table}[!htb]
    \centering
    \small
    \begin{tabular}{>{\tiny}l|c|c|c|c|c}
    \toprule
     & \multicolumn{5}{|c}{Conversation History Task} \\
     Target Task & AA & RT & TQ & GW & HA \\
     \midrule
     MMLU AA  & 	28.57 &	18.18 &	-4.76 &	-19.05 & *\\
     RT  & 	0.97 &	3.79 &	-0.87 &	-1.30 &	*  \\
     Tweet QA  & 	-0.78 &	7.62 &	30.56 &	9.26 & *\\
     Gigaword  & 	-3.81 &	2.61 &	3.44 &	78.71 &	* \\ 
     MMLU HA  & 	* & 1.63 &	0.81 &	-11.38 & 12.20 \\
     \bottomrule
    \end{tabular}
    \caption{Model: \textbf{Mistral-7B}. Percentage \% change in model performance.}
    \label{tab:mistral-confuse}
\end{table}

\section{Free-run Performance}
\label{sec:freerun}

\begin{figure*}[h]
\centering
\begin{subfigure}{\linewidth}
    \includegraphics[width=\linewidth]{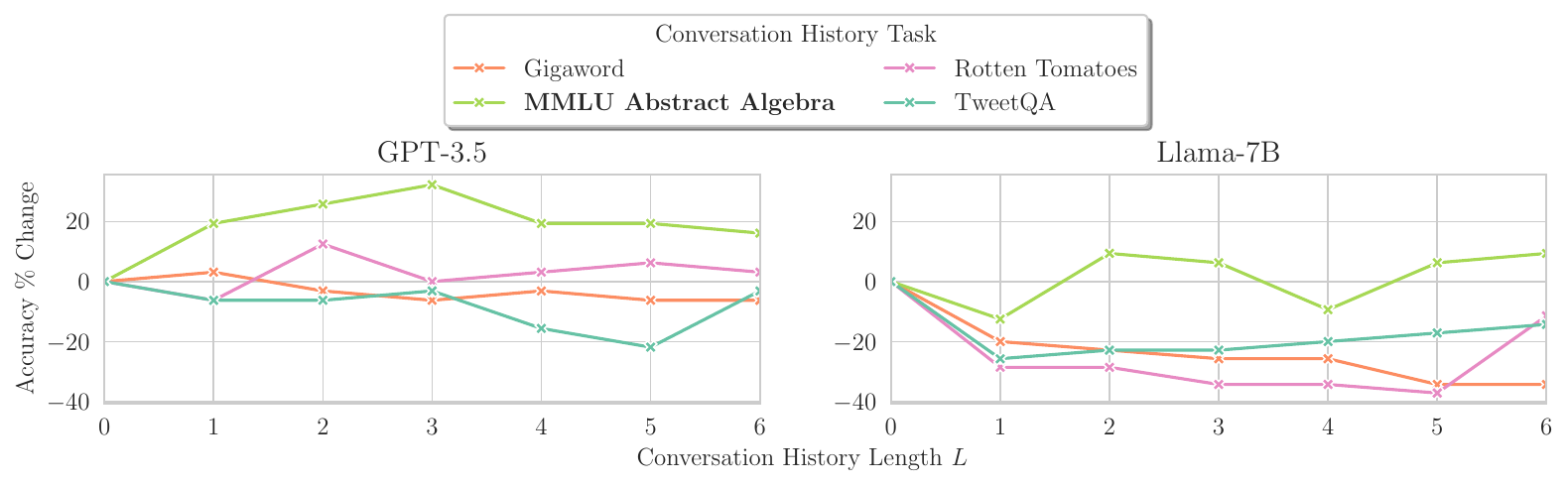}
    \label{fig:mmluaa-freerun}
\end{subfigure}
\begin{subfigure}{\linewidth}
    \includegraphics[width=\linewidth]{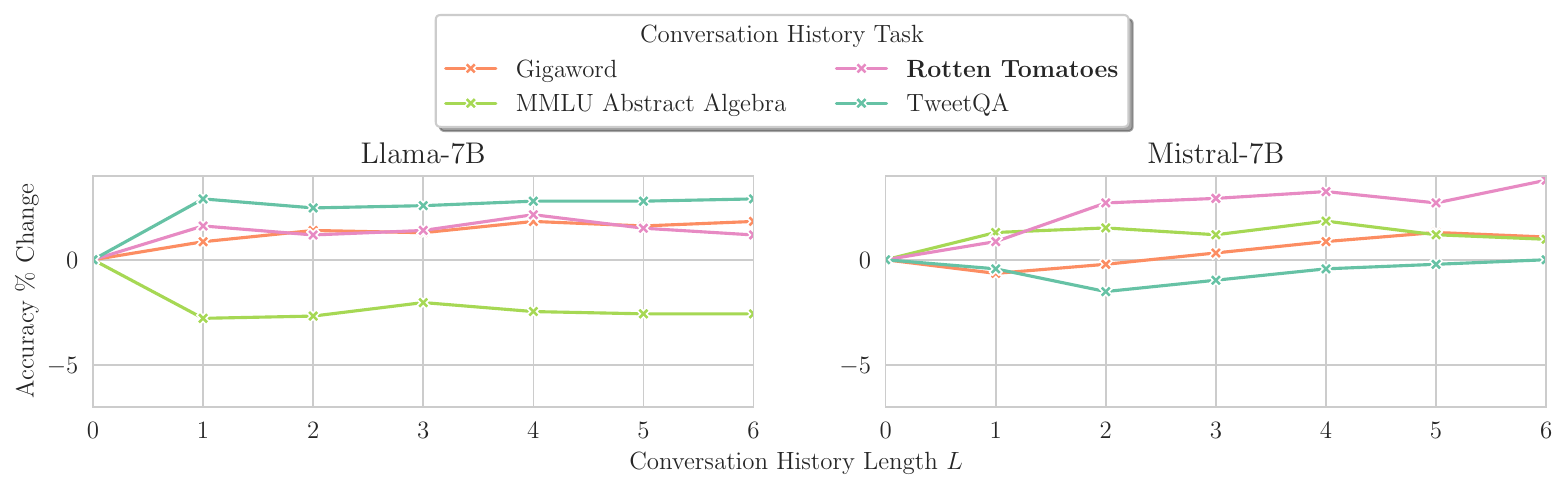}
    \label{fig:rt-freerun}
\end{subfigure}
\begin{subfigure}{\linewidth}
    \includegraphics[width=\linewidth]{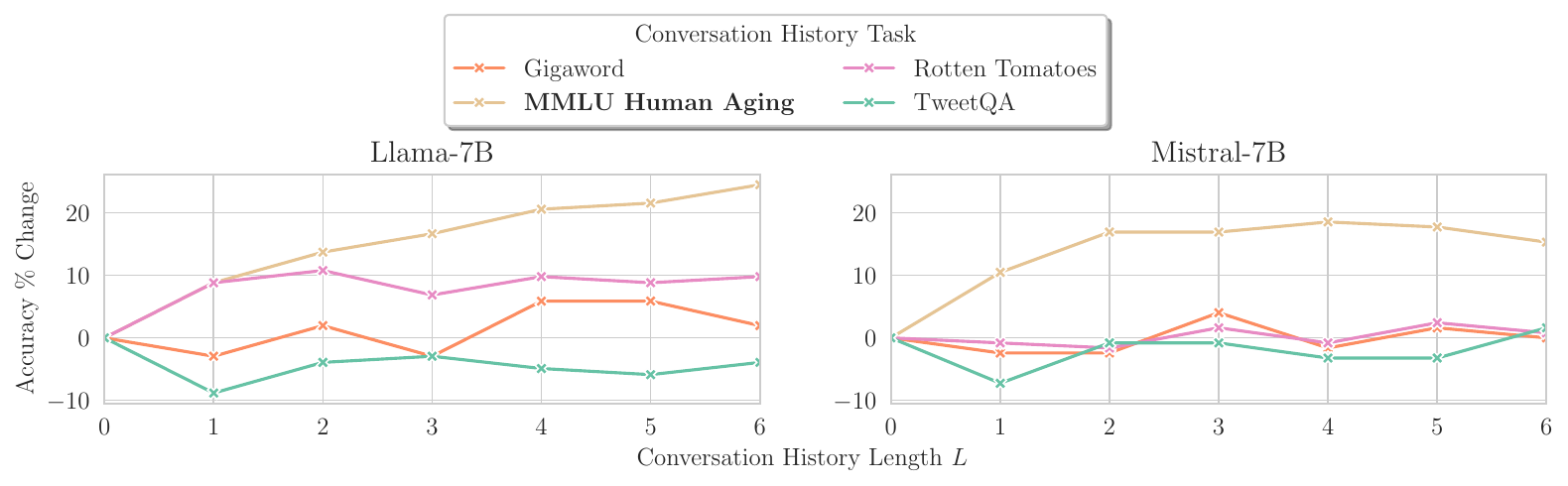}
    \label{fig:mmlu-age-freerun}
\end{subfigure}
\caption{Percentage \% change in accuracy \emph{without} teacher-forcing i.e. free-running, relative to zero-shot performance (no conversation history) for increasing conversation history length $L$ and various models. The target tasks are (MMLU Abstract Algebra, Rotten Tomatoes and MMLU Human Aging). }
\label{fig:freerun}
\end{figure*}

In this section, we investigate the performance change for task-switch without teacher-forcing the response, $r_k$, in the conversation history (mentioned in our methods at the end of Section~\ref{sec:method}). We call this ``free-running''. The results of the free-running experiments are shown in Fig~\ref{fig:freerun}. 

Similar to the teacher-forced experiments, we observe that the ``free-running'' experiments show that model performance changes in the presence of a task-switch. We note that the results are indeed not identical because these models are not perfect at responding to the history tasks, $T_h$. In Fig~\ref{fig:freerun}, we observe that the performance can degrade based on the combination of model and task-switch. Interestingly, some task-switches allow the model to perform better than in-context examples, evident in Llama-7B for Rotten Tomatoes as $T_t$ and Tweet QA as $T_h$.

\section{Random Conversation History}
\label{sec:rand_hist}

In this section, we study the impact on the task-switch performance with a \textit{random} conversation as the history, as opposed to a specific single conversation-history task, $T_h$. This evaluation offers a baseline measure of a model's general sensitivity to a change from the conversation history. This allows us to assess the extent to which the observed performance degradation from a task-switch is due to the explicit change in task, as opposed to the presence of an unrelated conversation history.

To perform this experiment we generate $K=20$ random conversation pairs $(u_k, r_k)$ as follows:

\begin{itemize}
    \item Model generates the random user utterance, $u_k$, starting from the \texttt{<BOS>} (beginning of sequence) token.
    \item Model generates the system response, $r_k$, in a conversational setting.
\end{itemize}
\noindent
This gives us a set of user-system conversation pairs, which we can use to create a `random' conversation history of length $L$ by sampling pairs from this dataset. 

\begin{figure*}[h]
    \centering
    \includegraphics[width=\linewidth]{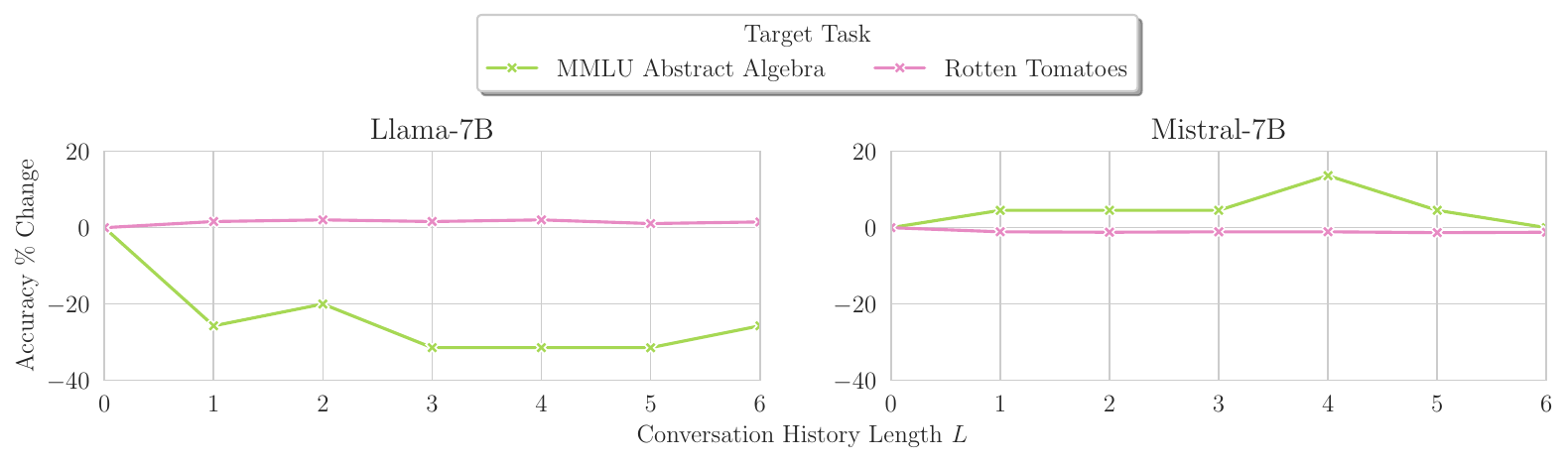}
    \caption{Percentage \% change in accuracy with a \underline{random conversation history} for increasing conversation history length $L$. Please note that in this Figure, the plotted lines correspond to the target task, not the conversation history task.}
    \label{fig:rand_hist}
\end{figure*}

\noindent
These experiments are evaluated on the models: Llama-7B and Mistral-7B, for the target tasks ($T_t$): MMLU Abstract Algebra (MMLU AA) and Rotten Tomatoes (RT). The results are shown in Fig~\ref{fig:rand_hist}. We observe that both models have no task-switch performance change with $T_t=$ RT. However, both models have some performance change with MMLU AA as $T_t$; noticeably, Llama-7B suffers the most. We hypothesize that these results can be explained by the task-switch sensitivity metric $\tau$ reported in Tables~\ref{tab:results-mmluaa},~\ref{tab:results-rt}. Table~\ref{tab:results-rt} shows a comparatively lower task-switch sensitivity compared to MMLU AA as $T_t$ in Table~\ref{tab:results-mmluaa}. Moreover, with MMLU AA as $T_t$ (Table~\ref{tab:results-mmluaa}), Llama-7B consistently has a higher task-switch sensitivity than Mistral-7B, which could explain the large performance change observed in Fig~\ref{fig:rand_hist}. Overall, these results confirm that observed performance degradation due to task-switches is primarily due to the change in task and not due to the presence of random/unrelated noise in the conversation history.

\end{document}